\documentclass{article}

\usepackage{graphicx} 
\usepackage{subfigure}
\usepackage{booktabs} 
\usepackage{multirow}
\usepackage{array}
\usepackage{rotating}
\usepackage{xcolor}
\usepackage{times}
\usepackage{soul}
\usepackage{url}
\usepackage[hidelinks]{hyperref}
\usepackage[utf8]{inputenc}
\usepackage{booktabs}
\usepackage{multirow}
\usepackage{algorithmic}
\usepackage{colortbl}
\usepackage[switch]{lineno}
\usepackage{float}
\usepackage[small]{caption}

\usepackage{hyperref}

\usepackage[accepted]{icml2025}

\usepackage{amsmath}
\usepackage{amssymb}
\usepackage{mathtools}
\usepackage{amsthm}
\usepackage{stmaryrd}  

\usepackage[capitalize,noabbrev]{cleveref}

\definecolor{orange}{RGB}{255,127,0}
\definecolor{brown}{RGB}{150,70,0}
\definecolor{green}{RGB}{127,255,127}
\definecolor{darkgreen}{RGB}{0,127,0}
\definecolor{blue}{RGB}{127,127,255}
\definecolor{lightblue}{RGB}{150,150,255}
\definecolor{darkblue}{RGB}{0,0,127}
\definecolor{red}{RGB}{255,90,90}
\definecolor{violet}{RGB}{200,110,170}
\definecolor{grey}{RGB}{127,127,127}
\definecolor{pink}{RGB}{255,180,180}

\newcommand{\LsgnN}{\ensuremath{L^{\mp}_1}} 
\newcommand{\LabsN}{\ensuremath{L^{+}_1}} 
\newcommand{\LsgnS}{\ensuremath{L^{\mp}_2}} 
\newcommand{\LabsS}{\ensuremath{L^{+}_2}} 
\newcommand{\LsgnL}{\ensuremath{L^{\mp}_L}} 
\newcommand{\LabsL}{\ensuremath{L^{+}_L}} 

\newcommand{\proxy}{\ensuremath{\circ\!\shortrightarrow}}
\newcommand{\target}{\ensuremath{\shortrightarrow}\!\circ}

\newcommand{\vect}[1]{\ensuremath{\mathbf{#1}}}

\newcommand{\model}[1]{{\small\textsf{#1}}}
\newcommand{\modelc}[1]{{\scriptsize\textsf{#1}}}

\newcommand{\ds}[1]{{\small\textsf{#1}}}
\newcommand{\dsc}[1]{{\scriptsize{}\textsf{#1}}}

\theoremstyle{plain}
\newtheorem{theorem}{Theorem}[section]

\theoremstyle{definition}
\newtheorem{definition}[theorem]{Definition}

\theoremstyle{remark}

\usepackage[textsize=tiny]{todonotes}

\icmltitlerunning{What should an AI assessor optimise for?}

\begin{document}

\twocolumn[
\icmltitle{What should an AI assessor optimise for?}



\icmlsetsymbol{equal}{*}

\begin{icmlauthorlist}
\icmlauthor{Daniel Romero-Alvarado}{upv}
\icmlauthor{Fernando Martínez-Plumed}{upv}
\icmlauthor{José Hernández-Orallo}{upv}
\end{icmlauthorlist}

\icmlaffiliation{upv}{VRAIN, Universitat Politècnica de València, Valencia, Spain}

\icmlcorrespondingauthor{Daniel Romero-Alvarado}{dromalv@vrain.upv.es}
\icmlcorrespondingauthor{Fernando Martínez-Plumed}{fmartinez@dsic.upv.es}
\icmlcorrespondingauthor{José Hernández-Orallo}{jorallo@upv.es}

\icmlkeywords{Machine Learning, evaluation, assessor models}

\vskip 0.3in
]



\printAffiliationsAndNotice{}  

\begin{abstract}
An AI assessor is an external, ideally independent system that predicts an indicator, e.g., a loss value, of another AI system. Assessors can leverage information from the test results of many other AI systems and have the flexibility of being trained on any loss function or scoring rule: from squared error 
to toxicity metrics. Here we address the question: is it always optimal to train the assessor for the target metric? Or could it be better to train for a different metric 
and then map predictions back to the target metric? Using 
twenty regression and classification problems with tabular data, we experimentally explore this question for, respectively, regression losses and classification scores with monotonic and nonmonotonic mappings and find that, contrary to intuition, optimising for more informative 
metrics is not generally better. Surprisingly, 
some monotonic transformations 
are promising. For example, the logistic loss is useful for minimising absolute or quadratic errors in regression, and the logarithmic score helps maximise quadratic or spherical scores in classification.
\end{abstract}

\section{Introduction}
\label{sect:introduction}

AI models and systems are evaluated with very different metrics, depending on the purpose of application. For instance, metrics as diverse as the BLEU score \cite{papineni2002bleu} for translation, `Bold' toxicity score \cite{dhamala2021bold} for text generation,  the area under the ROC curve \cite{fawcett2006introduction} 
for classification, asymmetric loss  \cite{elliott2005estimation} 
for sales prediction \cite{gogolev2023asymmetric} or any reward function \cite{eschmann2021reward} for reinforcement learning, are commonly used. 
Models can be built or trained to minimise some loss, and then repurposed for a situation where another metric matters more. The most characteristic example today of this process is represented by `foundation models' \cite{bommasani2021opportunities}, such as language models. Even if the model can produce uncertainty estimates about the next token, and these are well calibrated, the metric of interest may be toxicity, for example. 
Since the model does not estimate toxicity, we need some external way to do this.

\begin{figure}[t]
    \centering
    \includegraphics[width=0.9\linewidth]{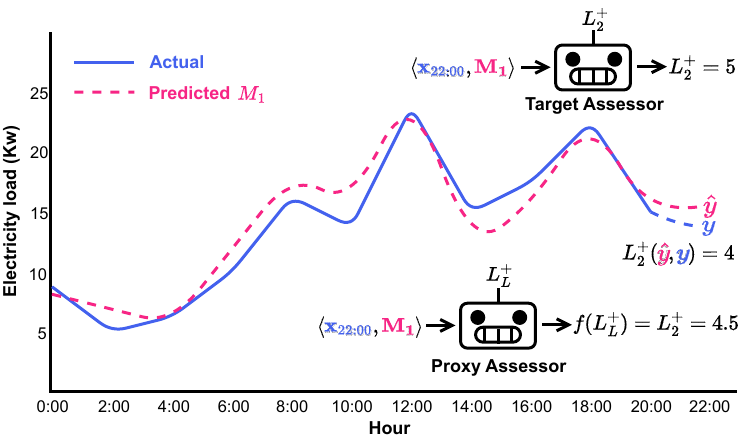}
    \caption{For an energy consumption model $M_1$, we want to anticipate the squared error ($\LabsS$) for each new example using an external predictor, called \textit{assessor}. Recommendations to customers are only made when the assessor predicts low $\LabsS$ 
    in the energy consumption estimate. 
    We will explore assessors that optimise for the target loss function (squared loss $\LabsS$, top) but also assessors that use a proxy loss function (logistic loss $\LabsL$, bottom) followed by a transformation ($f$). Can the proxy assessor be better?}
    \label{fig:example}
\end{figure}

One solution to this challenge is the development of \textit{assessor models} \cite{hernandez2022training}. 
An assessor 
is a predictive model designed to estimate how well another system, called the base or subject system $s$, will perform on a given example or problem instance $i$ 
for a specific validity metric before it is actually deployed. An assessor 
can estimate the conditional distribution $\hat{p}(v|s,i)$ or simply (pointwise) map $\langle s,i\rangle \mapsto v$. Assessors 
are related to verifiers \cite{li2023making} but are \textit{anticipatory}: rather than simply checking outcomes post-execution, they predict the outcomes in advance (i.e.,  
given a new example $i$, they can predict the value $v$ of the metric that $s$ is expected to achieve). For instance, consider $s$ a self-driving car and $i$ a specific journey. An assessor could predict the safety outcome $v$ of $s$ for $i$. 

Assessors are used to anticipate any metric of quality, safety, bias or, in general, validity for any kind of subject system, from RL agents to language models. Assessors can be used to monitor or forecast system performance \cite{schellaert2024proposal}, to optimise configurations \cite{zhao2024team}, to do anticipatory reject \cite{zhou2022reject}, 
or to delegate by routing \cite{hu2024routerbench,lu2023routing,ding2024hybrid}. 
Assessors are usually trained on test data, capitalising on vast information from results of many systems and examples \cite{burnell2023rethink}.

It may seem natural that the assessor is trained to optimise for the metric we are interested in. For instance, if the subject system $s$ estimates daily energy consumption of households and the metric value $v$ is given by the squared error (\LabsS) between actual and estimated consumption values, then one would expect that the assessor should be trained to predict the squared error that the system will incur for each household. 
However, in this paper \textit{we challenge the general assumption that training an assessor to optimise directly for a specific metric $L$ necessarily results in the best optimisation outcome for $L$}. In this example, what if optimising for logistic loss ($\LabsL$) were better? This situation is illustrated in  Figure~\ref{fig:example}.

To start exploring this question, in this paper we will consider the base model is solving a regression problem and we will use generic regression metrics, such as absolute error, squared error and logistic error. We will consider signed and unsigned (absolute) versions of these three metrics, and explore whether optimising for a proxy metric is better than optimising for the target metric. From our experimental analysis we observe some results that may be explained by the distribution of errors (residuals) in the test data of the base subject systems. However, some other results are more surprising, such as the logistic error being the best in all situations. Give the effectiveness of some monotonic transformations for regression, we also explore the case of classification, focusing on monotonic transformations, and find a similar phenomenon. These findings suggest that learning an assessor for one central metric might suffice to optimise a family of monotonically-related metrics.

\section{Background}
\label{sect:background}

This work situates itself within a broad spectrum of research on error analysis 
and the exploration of alternative loss functions for training predictive models. However, the use of assessors resituates this question at the meta-level, as a second-order regression problem, an area that, to our knowledge, has not been explored yet. 

\subsection{Error analysis in regression} 
In regression problems, the choice and optimisation of loss functions is critical to model performance. There is an extensive literature on traditional error measures \cite{hyndman2006another,botchkarev2018performance,botchkarev2019new,chicco2021coefficient} such as Mean Squared Error (MSE), Absolute Error, and more robust variants such as Huber Loss \cite{owen2007robust}, 
which falls somewhat in between squared and absolute error, or Tukey's biweight loss \cite{beaton1974fitting,belagiannis2015robust}, which caps quadratic loss beyond a given point. 
Optimisation of these loss functions leads to different kinds of bias. For instance, quadratic error leads to estimators that are unbiased for the mean while absolute error leads to estimators that are unbiased for the median. 


 
Beyond their use in performance evaluation, the analysis of errors and residuals also serves a diagnostic purpose, helping to identify model inadequacies or violations of assumptions, providing a comprehensive understanding of the linear and non-linear relationships captured by regression models. For instance, \cite{rousseeuw2005robust}
use regression diagnostics, e.g.,  outlier diagnosis, to identify problems in both the explanatory and response variable, 
further refining the understanding of errors in predictive models.


Some studies have also explored more complex loss functions and their impact on regression model performance. According to \cite{gneiting2007strictly}, appropriate scoring rules incentivise truthful prediction by optimising prediction distributions. 
However, as models and tasks become more complex, optimising a single loss function  
may not always align 
with the broader objectives of the system.
In this regard, 
research such as \cite{
huber1992robust} 
experiment with alternative, often non-convex, loss functions designed to improve model training under specific constraints or performance benchmarks. 

\subsection{Error analysis in classification}

In probabilistic classification, evaluating predicted probabilities is as crucial as evaluating label accuracy. Proper scoring rules~\cite{gneiting2007strictly} such as the log score (cross entropy), Brier score (quadratic score), and spherical score assess probabilistic forecasts' accuracy and promote honest reporting of uncertainties when classifiers provide probability distributions. 


The choice of loss functions used during training also affects classification errors. Cross-entropy loss (or log loss) is commonly used to train neural networks for classification tasks because it is consistent with maximising the probability of the correct class~\cite{bishop2006pattern}. Alternative loss functions, such as the hinge loss used in support vector machines~\cite{cortes1995support}, have different properties and can lead to different error patterns.

In this paper we explore proper scoring rules, as they are minimised when converging to the true distribution and, more importantly, all of them can be expressed as a a distribution between $[0,1]$ of all decision threshold (e.g., Brier score assumes a uniform distribution on thresholds)\cite{buja2005loss,hernandez2012unified}. So it makes sense to explore first if some of them can be used as proxy for the others.

\subsection{Assessors}  
%
The concept of \textit{assessors} was first introduced in \cite{hernandez2022training}. \cite{kadavath2022LanguageModelsMostly} extended this by examining LLM and their role as assessors, finding that larger models tended to be more accurate and consistent in predicting outcomes across multiple tasks, although they acknowledged a lack of generalisation in out-of-distribution scenarios. \cite{schellaert2024analysing} further explored assessors to predict instance-level LLM performance on over 100 BIG-bench \cite{srivastava2022bigbench} tasks \cite{schellaert2024analysing}, outperforming subject systems in confidence and demonstrating scalability across model sizes. Similarly, \cite{zhou2022reject} showed that smaller LLMs can predict the performance of larger models on certain tasks, significantly reducing errors and computational costs. \cite{pacchiardi2024100instancesneedpredicting} also proposed estimating LLM performance using a limited set of reference instances.
Other applications of assessors focus on forecasting system performance (scaling laws) \cite{schellaert2024proposal}, team configurations \cite{zhao2024team}, anticipatory reject \cite{zhou2022reject},
or delegation (routing) to the best language model depending on the prompt 
\cite{lu2023routing,hu2024routerbench,ding2024hybrid}.  
However, an analysis of the chosen validity metric and its distribution has not been done to date.

An assessor is an 
external, second-order system 
that predicts the scores of another, first-order system, the subject. It is \emph{populational}, trained on test data spanning numerous instances and potentially multiple subjects. It operates as a \emph{standalone} entity, independent of the subject. This attribute allows it to be \emph{anticipatory}; it can predict the subject's performance solely on the basis of the input and the subject's characteristics, without needing access to the subject's output or the ability to execute it. Furthermore, the {\em standalone} nature of assessors offers advantages in terms of accountability and verification, as they can be developed by external auditors or for datasets different from those used to train the original subject. In addition, their use extends to increasing curriculum complexity, as in \cite{bronstein2022EmbeddingSyntheticOffPolicya}, or facilitating instance-level model selection, a concept derived from algorithm selection \cite{kerschke2019AutomatedAlgorithmSelection}. Finally, a perfect assessor (in an ideal scenario) would completely capture the epistemic uncertainty (error) associated with the subject's performance \cite{hullermeier2021aleatoric}, with the error of the assessor depending only on the aleatoric error of the subject. 

\begin{figure}[t]
    \centering
    \includegraphics[width=0.6\linewidth]{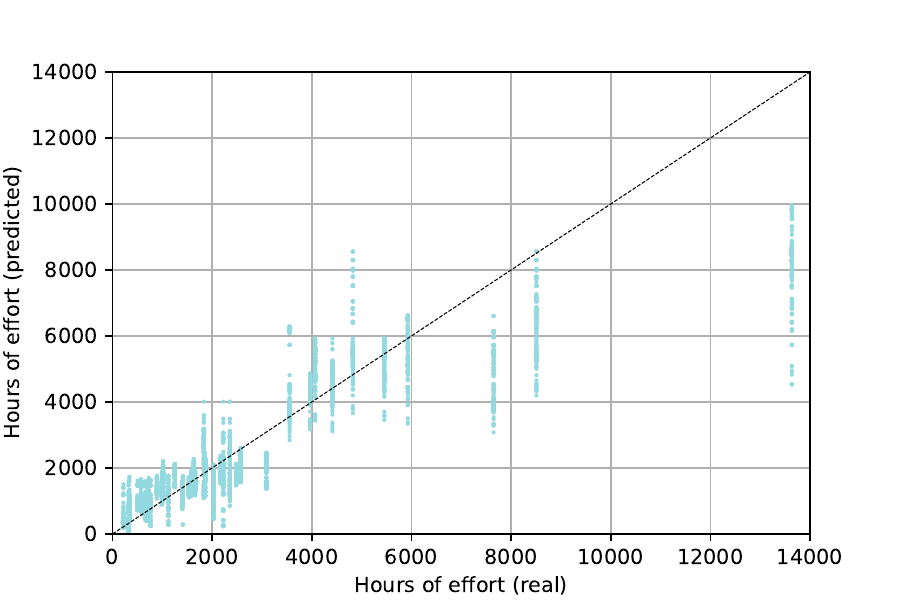}
    \includegraphics[width=\linewidth]{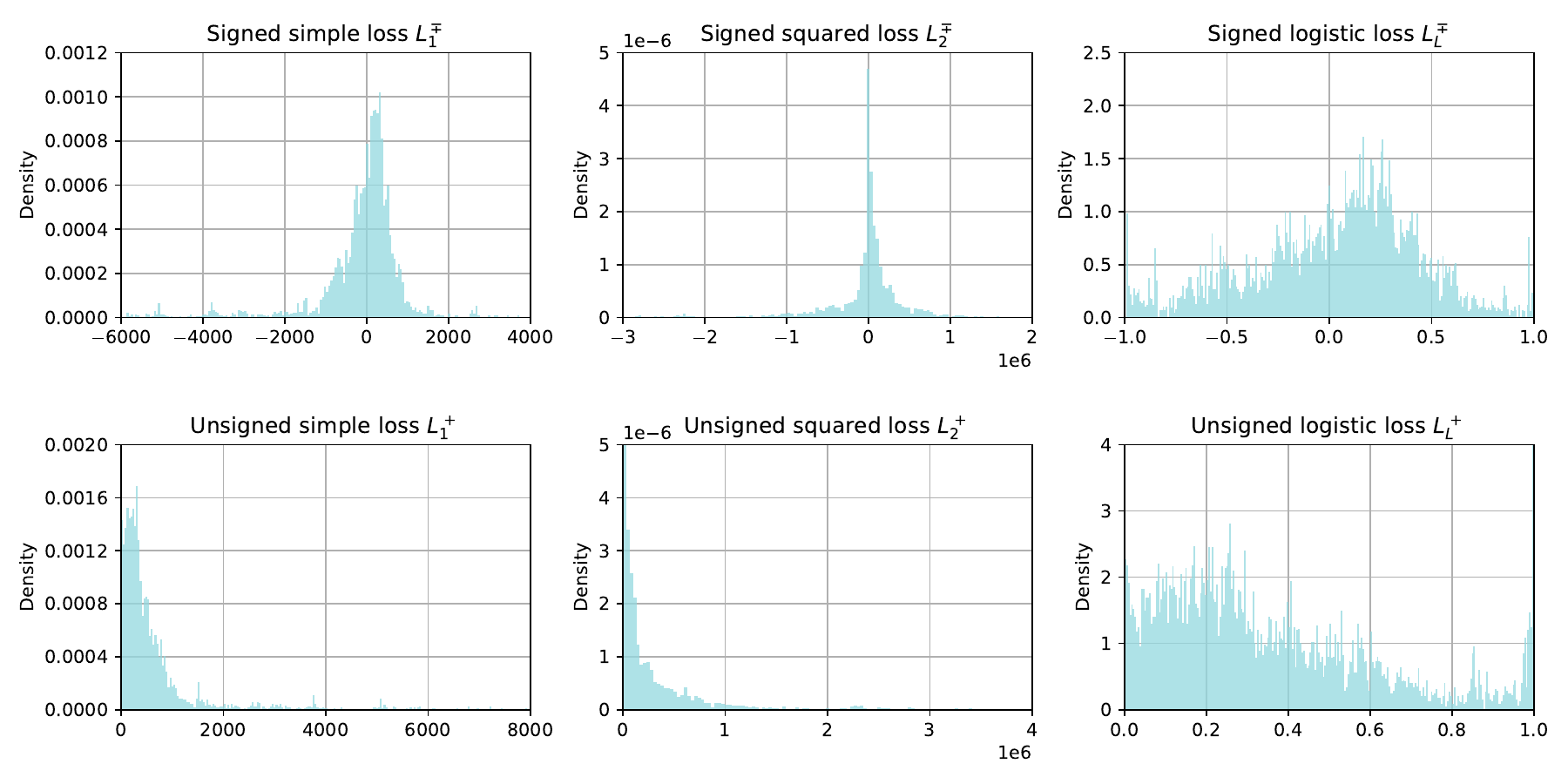}
    \caption{\dsc{Software Effort} dataset (Table~\ref{tab:ds}) 255 regression models. Top: scatter plot of $\hat{y}$ versus $y$. Bottom: histogram of losses: simple ($\LabsN$), squared ($\LabsS$), and logistic ($\LabsL$). 
    The top and bottom rows show the signed and unsigned versions. Assessors predict these losses, noting differences in shape and tails.
    }
    \label{fig:base}
\end{figure}

\begin{figure}[h]
    \centering
    \includegraphics[width=0.36\linewidth, height=90pt]{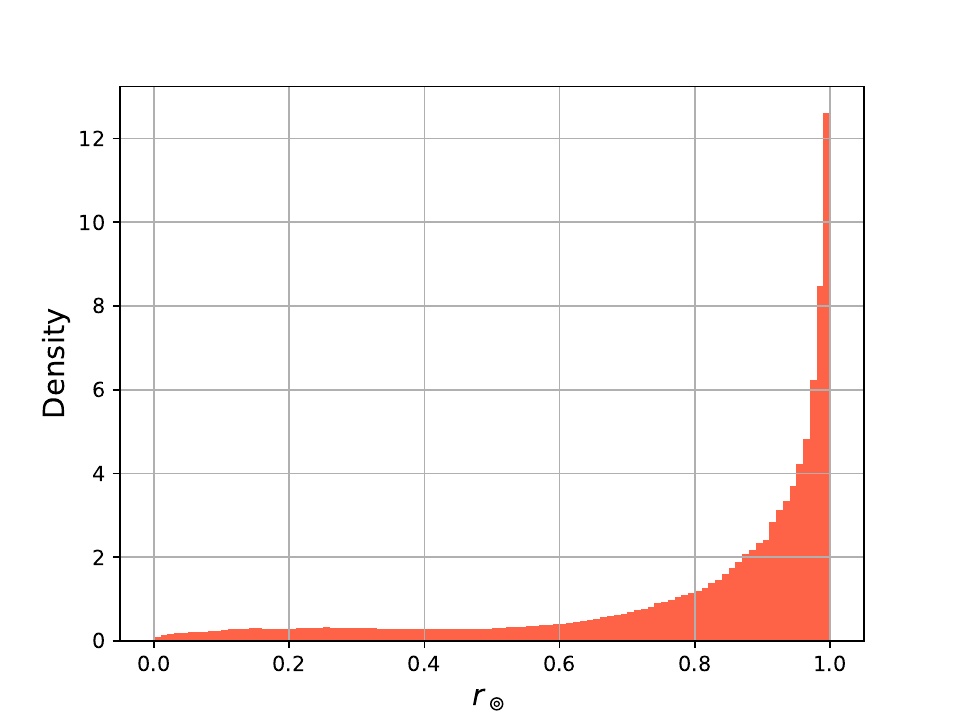}\hspace{-0.44cm}
    \includegraphics[width=0.36\linewidth, height=90pt]{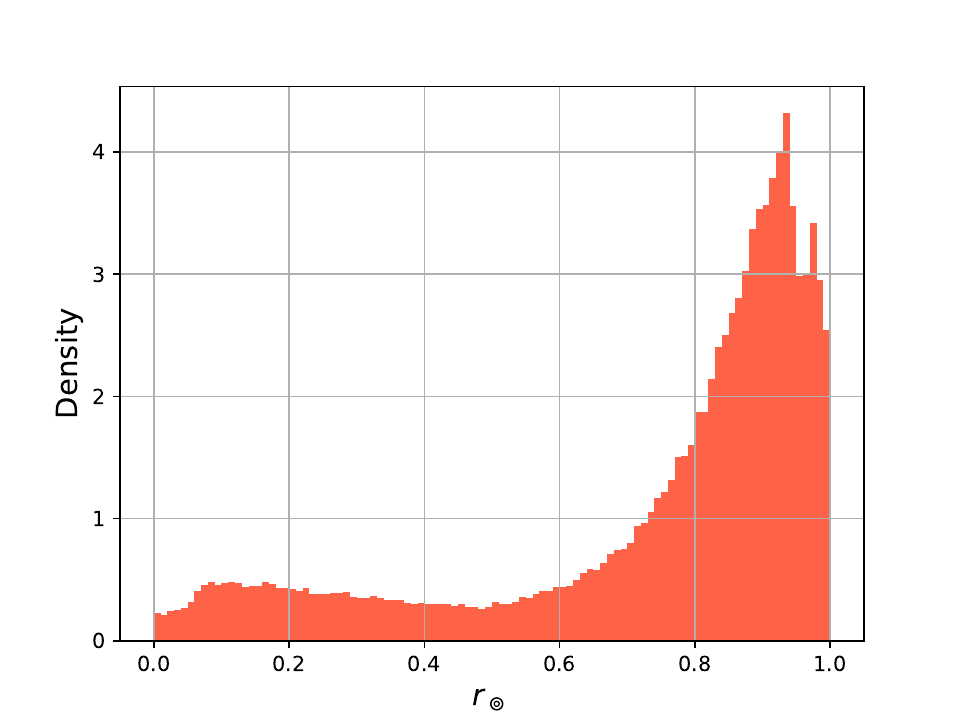}\hspace{-0.44cm}
    \includegraphics[width=0.36\linewidth, height=90pt]{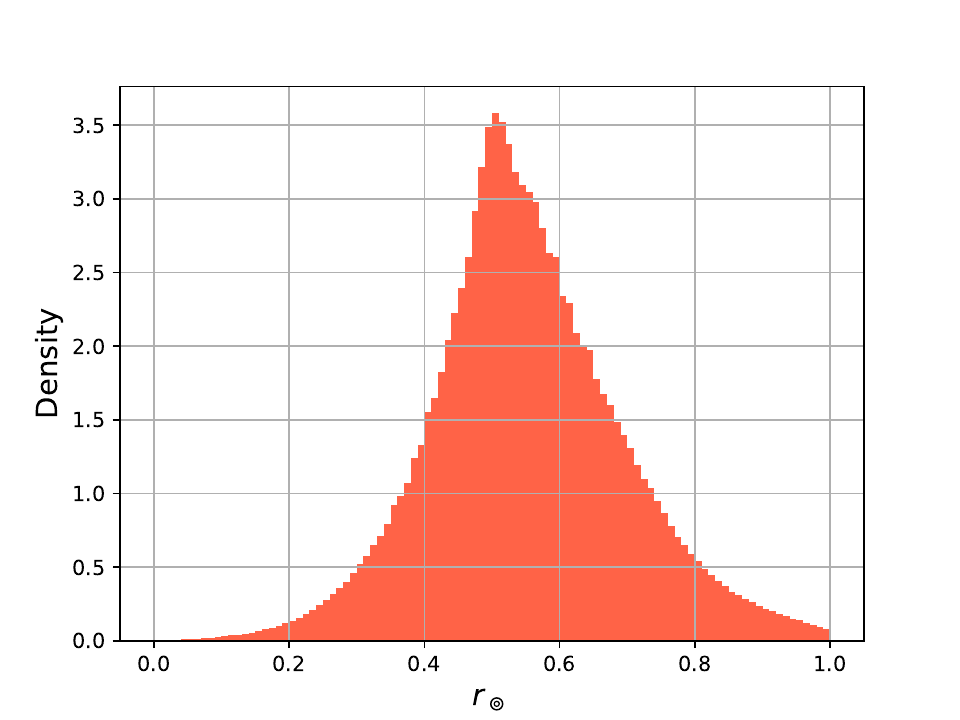}
    
    \caption{Histograms of $r_\circledcirc$, the probability estimation for the correct class for some representative datasets, all averaging 255 classification models:  \dsc{CDC Diabetes} (most have this shape), \dsc{JM1} (slightly bimodal) and \dsc{Higgs} (quite bad with values around 0.5). All datasets in Appendix~\ref{app:histograms}.}
    \label{fig:base_r_target}
\end{figure}


Assessors must learn from a very specific kind of distribution, given by the results of a loss function (or scoring rule) applied to the predictions of the base model. For instance, if this loss function is based on residuals, the dependent variable in the regression problem the assessors have to deal with will be affected by the distribution of residuals. Depending on the base model, this distribution may be normal or asymmetric, but the outliers tend to be of aleatoric character rather than epistemic. To illustrate this, Figure~\ref{fig:base} (top) shows a scatter plot for the predicted and actual values of the 
Software Effort 
test set with 255 regression models.
We see some outliers near 14000 for which models predict values between 4000 and 10000, leading to high residuals. This suggests that giving lower proportional weight to these errors in the loss function, as the $L_L$ loss in the bottom image does, may be a particularly interesting route to explore for assessors. This hypothesis is behind the experimental methodology in the following section.

Similarly, in classification problems, the choice of scoring rule used by the assessor affects the distribution of the dependent variable that it must predict~\cite{gneiting2007strictly}. For instance, the logarithmic score heavily penalises confident but incorrect predictions, which can lead to a distribution with pronounced tails where misclassifications with high confidence dominate the learning signal and are hence heavily penalised when wrong. The quadratic score, on the other hand, provides a smoother gradient. 
Figure~\ref{fig:base_r_target} shows the histogram of the estimated probability of the correct class $r_\circledcirc$ from 255 classification models on a subset of datasets from~\ref{tab:ds}. Few cases have 
probabilities below 0.2, meaning that high confidence misclassifications are very rare. Using the logarithmic score for assessors may enhance its sensitivity to these significant misclassifications. The \ds{Higgs}  dataset (Figure~\ref{fig:base_r_target} (right)) is an exception exhibiting a symmetrical distribution of observed class probabilities. Here,  the logarithmic score's heavy penalisation of low-probability correct classes will not work that well.

\section{Metrics and Problem Representation}
\label{sect:problem_rep}

For the rest of the paper, base subjects $m_s$ are 
models $m_s : X \mapsto Y$, where $X \subset \mathbb{R}^d$ is an input feature vector and $Y \subset \mathbb{R}$ or $Y \subset \{0, 1\}$ is the output, depending on whether the task is regression or binary classification. For \textbf{regression}, given the output $\hat{y} = m_s(\vect{x})$ and the ground truth $y$, we can calculate any metric or loss function $L: \mathbb{R} \times \mathbb{R} \mapsto \mathbb{R}$, denoted as $L(\hat{y},y)$. We will consider the following {\em signed} loss functions, all as a function of the residual $e = \hat{y} - y$:

\begin{definition}Signed simple error
\begin{equation}
\LsgnN(\hat{y},y) := \hat{y}-y
\end{equation}
\end{definition}

\begin{definition}Signed squared error
\begin{equation}
\LsgnS(\hat{y},y) := 
(\hat{y}-y) \cdot |\hat{y}-y|
\end{equation}
\end{definition}

\begin{definition}Signed logistic error
\begin{equation}
\LsgnL(\hat{y},y) := \frac{2}{1 + e^{-B(\hat{y}-y)}} - 1, B = \frac{\ln{3}}{\mathrm{mean}_{Y}{|\hat{y}-y|}} 
\end{equation}
\end{definition}

\noindent 
The signed logistic error is a derivation from the general formula for a logistic curve 
so that values near $-1$ correspond to high underpredictions and values near $1$ correspond to high overpredictions. Additionally, since different regression tasks can have different ranges of errors (for instance, errors when predicting the number of rings in trees do not have the same magnitude as errors when predicting house pricings), we parametrise $\LsgnL$ by a value $B$, such that the value of $\LsgnL$ is 0.5 when the error in an instance is equal to the mean of the absolute errors of the base model. 

The corresponding unsigned loss functions, are defined by simply removing the sign, i.e., $\LabsN := |\LsgnN|$, $\LabsS := |\LsgnS|$ and $\LabsL := |\LsgnL|$. It is easy to see that $\LsgnN$, $\LsgnS$ and $\LsgnL$ are mononotically related (they do not lose information between each other), and the same happens between the unsigned versions. Of course, this no longer happens between the signed and unsigned versions, as the unsigned versions lose information. Figure~\ref{fig:losses} in 
 the Appendix shows the six losses. The signed losses contain information about the \textit{magnitude} and the \textit{direction} of the error, whereas their unsigned counterparts only carry the \textit{magnitude}, hence being less informative. 
The logistic loss tries to represent a smooth loss function that penalises outliers (mostly of aleatoric character) proportionally less than lower errors. It is hence a non-convex loss that, 
unlike the Huber Loss, 
does not fall in between the simple (linear) and squared errors, but goes beyond the linear error. It saturates on high residuals, but unlike Tukey's biweight loss, it is not piecewise, and has non-zero gradient everywhere (Tukey's loss is constant from a value, which is usually chosen to be 4.685 when residuals follow a standard normal distribution) \cite{belagiannis2015robust}.

For \textbf{classification}, given a forecasted probability vector $m_s(\vect{x}) = \vect{r}$, $m$ different classes and ${r_\circledcirc}$ as the probability assigned by the model to the observed 
class, known as the `principal', we will consider the following proper scoring rules $S: [0,1]^m \mapsto \mathbb{R}$ (see Figure~\ref{fig:scoringrules} their  functional representation):

\begin{definition}Logarithmic score
    \begin{equation}
        S_L(\vect{r}) := ln({r_\circledcirc})
    \end{equation}
\end{definition}
\begin{definition}Quadratic score
    \begin{equation}
        S_Q(\vect{r}) := 2{r_\circledcirc} - \vect{r} \cdot \vect{r}
    \end{equation}
\end{definition}
\begin{definition}Spherical score
    \begin{equation}
        S_S(\vect{r}) := \frac{{r_\circledcirc}}{||\vect{r}||}
    \end{equation}
\end{definition}

In the case of binary classification, we can define $\vect{r}$ as a two-component vector only in terms of the principal ${r_\circledcirc}$
 , simplifying some of the previous scores, we get:

{\small{
\begin{minipage}[h]{0.4\columnwidth}
\begin{align}
    S_Q(\vect{r}) &= -(2{r_\circledcirc}^2 - 4{r_\circledcirc} + 1) 
\end{align}
\end{minipage}
\hspace{0.075\columnwidth}
$\:\:\:\:\:\:$
\begin{minipage}[h]{0.4\columnwidth}
\begin{align}
  S_S(\vect{r}) &= \frac{{r_\circledcirc}}{\sqrt{2{r_\circledcirc}^2 - 2{r_\circledcirc} + 1}} 
\end{align}
\end{minipage}
}}

Which now allows for transformations between them:

{\scriptsize{
\begin{minipage}[h]{0.4\columnwidth}
\begin{align}
      S_S &= \frac{e^{S_L}}{\sqrt{2e^{2S_L} - 2e^{S_L} + 1}} \\
    S_Q &= -(2e^{2S_L} - 4e^{S_L} + 1) \\
    S_Q &= \frac{4\left({S_S}^2 - \sqrt{{S_S}^2 - {S_S}^4}\right)}{2{S_S}^2 - 1} -\nonumber \\
        & \frac{2\left({S_S}^2 - \sqrt{{S_S}^2 - {S_S}^4}\right)^2}{(2{S_S}^2 - 1)^2} - 1
\end{align}
\end{minipage}
$\:\:\:$
\begin{minipage}[h]{0.4\columnwidth}
\begin{align}
   S_S &= \frac{2 - \sqrt{2 - 2S_Q}}{2\sqrt{2 - S_Q - \sqrt{2 - 2S_Q}}} \\
    S_L &= \ln\left(1 - \frac{\sqrt{2 - 2S_Q}}{2}\right) \\
    S_L &= \ln\left(\frac{{S_S}^2 - \sqrt{{S_S}^2 - {S_S}^4}}{2{S_S}^2 - 1}\right)
\end{align}
\end{minipage}
}}

A detailed derivation of these transformations is provided in Appendix~\ref{app:lossscoreexploration}. Note that the principal ${r_\circledcirc}$ plays a role analogous to the residual in regression: values of ${r_\circledcirc}$ close to $1$ indicate that the correct class has been assigned a high probability (small error or residual), whereas values close to $0$ indicate that the correct class has been assigned a low probability (large error or residual). 

Once we have defined the loss and score functions, we need to 
formalise 
how to train assessors properly. 
Consider a class of subject systems $M$, which are represented by their size, number of parameters and other features, making a subject feature vector $\vect{m} \in M$. All these subject systems have previously been evaluated using a loss/score metric $L$. In order to build an assessor $a$, we need the input feature space $X$ and the subject space $M$ as inputs, and the metric as output, namely: $a: X \times M \mapsto \mathbb{R}$. The training set for the assessor is then composed of rows such as $\langle \vect{x}_i,\vect{m_s},l_{i,s}\rangle$, where $i$ and $s$ are the instance and system indexes respectively, $l_{i,s}=L(\hat{y}_{i,s},y_i)$ is the value to predict, with $y_i$ being the ground truth output for instance $i$, represented by $\vect{x}_i$ and $\hat{y}_{i,s} = m_s(\vect{x}_i)$.  
For classification, $l_{i,s}$ is replaced by one of the score functions $s_{i,s} = S(\vect{r}_{i,s})$.

In usual circumstances, $L$ is the {\em target} loss we care about and the one that appears in the training dataset for the assessor. However, in this paper we are going to distinguish between the target loss and the {\em proxy} loss. Consider that we build the training set $D_{tr}$ for the assessor with a proxy loss $L_{\proxy}$, and we train the assessor $a$ for this loss. If the target loss, $L_{\target}$, is different from the proxy loss, then we need to transform the output of the assessor $\hat{l}$ back to the target loss by using a transformation function $f$. 
This gives us two possible routes given a target loss $L_{\target}$: we can either train an assessor that directly optimises for $L_{\target}$ or train an assessor that optimises for a proxy error $L_{\proxy}$ and then transform the assessor predictions via $f$. For instance, the transformation function $f$ between the unsigned simple error and the unsigned squared error is $f(l)=l^2$. Following the example of energy consumption from Figure~\ref{fig:example}, we could train an assessor model to predict the target loss (squared error) or train an assessor to predict a proxy loss (such as the unsigned logistic loss $\LabsL$) and then transform the output to obtain $\LabsS$. The same logic follows for the proper scoring rules in the context of classification.

\section{Methodology and experimental setup}
\label{sect:methods}

\paragraph{Generation of base model results} Training an assessor for a specific task requires \textit{test results} from one or more base models. We generate these results using the procedure illustrated in Figure~\ref{fig:assessorevaldata} in Appendix~\ref{app:asstraining} and described formally below. The more data and models we have the more the assessor can generalise. The quality of the assessor would also depend on the parametrisation of $\vect{x}$ and $\vect{s}$. 
In this regard, we 
have built a 
collection of base models as a training resource for 
the assessor. We used 10 regression and 10 classification datasets of varying number of instances and attributes (see Table~\ref{tab:ds}), as well as different distributions of the target variable. 
We use different \textit{model configurations} (i.e., representing the combination of a model and its associated hyperparameters). Training such a model configuration on a specific dataset 
provides us instance-level results of the predicted and actual values on the test set, as well as additional metrics including training and inference time, and memory usage. 
These characteristics, paired with different hyperparameter values, define the model parametrisation $\vect{s}$.

\newcommand{\STAB}[1]{\begin{tabular}{@{}c@{}}#1\end{tabular}}
\renewcommand{\arraystretch}{0.3}
\begin{table}[t]
\caption{Summary of regression (top) and classification (bottom) datasets: number of features (\#Feat.) and instances (\#Inst.), the type of features 
(categorical or numerical) and their domain.}\label{tab:ds}
\setlength{\tabcolsep}{3pt}
\centering
\vspace{2pt}
\resizebox{\columnwidth}{!}{%
\begin{tabular}{@{}cp{8.1cm}ccccc@{}}
\toprule
& \multirow{ 2}{*}{\textbf{Dataset}}   & \multirow{ 2}{*}{\textbf{\#Feat.}}   & \multirow{ 2}{*}{\textbf{\#Inst.}}   & \multicolumn{2}{c}{\textbf{Feat. Types}}   & \multirow{ 2}{*}{\textbf{Domain}} \\ 
\cmidrule(lr){5-6}
                                                             &      &        &      & \textbf{Cat.}   & \textbf{Num.}   &    \\ 
\midrule
\arrayrulecolor[gray]{0.8}

\multirow{38}{*}{\STAB{\rotatebox[origin=c]{90}{\textbf{Regression}}}}

& \ds{Abalone} \cite{misc_abalone_1}                                & 8    & 4177   & $\bullet$   & $\bullet$   & Biology    \\ 
\cmidrule(lr){2-7}
& \ds{Auction Verification} \cite{AuctionVer}                       & 8    & 2043   & $\bullet$   & $\bullet$   & Commerce   \\ 
\cmidrule(lr){2-7}
& \ds{BNG EchoMonts} \cite{romano2021pmlb}                          & 10   & 17496  & $\bullet$   & $\bullet$   & Health     \\ 
\cmidrule(lr){2-7}
& \ds{California Housing} \cite{CalHousing}                         & 8    & 20640  & $\bullet$   & $\bullet$   & Real State \\ 
\cmidrule(lr){2-7}
& \ds{Infrared Thermography Temp.} \cite{InfraredData}              & 3    & 1020   & $\bullet$   & $\bullet$   & Health     \\ 
\cmidrule(lr){2-7}
& \ds{Life Expectancy} \cite{WHOLE}                                 & 21   & 2938   & $\bullet$   & $\bullet$   & Health     \\ 
\cmidrule(lr){2-7}
& \ds{Music Popularity} \cite{yashrajkakkad_music}                  & 14   & 43597  & $\bullet$   & $\bullet$   & Music      \\ 
\cmidrule(lr){2-7}
& \ds{Parkinsons Telemonitoring (\textit{motor})} \cite{Parkinsons} & 20   & 5875   & $\circ$     & $\bullet$   & Health     \\ 
\cmidrule(lr){2-7}
& \ds{Parkinsons Telemonitoring (\textit{total})} \cite{Parkinsons} & 20   & 5875   & $\circ$     & $\bullet$   & Health     \\ 
\cmidrule(lr){2-7}
& \ds{Software Cost Estimation} \cite{OralloswCSC}                  & 6    & 145    & $\bullet$   & $\bullet$   & Projects   \\ 

\midrule
\multirow{38}{*}{\STAB{\rotatebox[origin=c]{90}{\textbf{Classification}}}}
& \ds{Human} \cite{adult_2}                                & 14    & 48842   & $\bullet$   & $\bullet$   & Finance    \\ 
\cmidrule(lr){2-7}
& \ds{Bank Marketing} \cite{moro2014data}                       & 17    & 45211   & $\bullet$   & $\bullet$   & Finance   \\ 
\cmidrule(lr){2-7}
& \ds{Magic gamma Telescope} \cite{magic_gamma_telescope_159}                          & 12   & 19020  & $\circ$   & $\bullet$   & Health     \\ 
\cmidrule(lr){2-7}
& \ds{JM1} \cite{menzies2004good}                         & 22    & 10885  & $\circ$   & $\bullet$   & Software \\ 
\cmidrule(lr){2-7}
& \ds{Nomao} \cite{candillier2012design}              & 119    & 34465   & $\circ$   & $\bullet$   & Entities     \\ 
\cmidrule(lr){2-7}
& \ds{Pen Digits} \cite{pendigits}                                 & 16   & 10992   & $\circ$  & $\bullet$   & CS     \\ 
\cmidrule(lr){2-7}
& \ds{CDC Diabetes} \cite{burrows2017incidence}                  & 21   & 253680  & $\bullet$   & $\bullet$   & Health      \\ 
\cmidrule(lr){2-7}
& \ds{Higgs} \cite{higgs_280} & 28   & 500000   & $\circ$     & $\bullet$   & Physics     \\ 
\cmidrule(lr){2-7}
& \ds{Credit Card Fraud} \cite{dal2014learned} & 30   & 284808   & $\circ$     & $\bullet$   &  Finance     \\ 
\cmidrule(lr){2-7}
& \ds{R\&K vs K Checkmates} \cite{chess_23}                  & 6    & 28056    & $\circ$   & $\bullet$   & Chess   \\ 
\arrayrulecolor{black}
\bottomrule
\end{tabular}
}
\end{table}

In order to have a homogeneous parametrisation $\vect{s}$ we train five distinct tree-based algorithms for each of the 
twenty 
datasets. 
Our base models include \model{Decision Trees}~\cite{BreimanDT}, \model{Random Forests}~\cite{HoRF}, \model{CatBoost}~\cite{catboost}, \model{XGBoost}~\cite{XGB}, and \model{LightGBM}~\cite{lightGBM}. For each dataset, we generate 255 different unique model configurations (denoted by the system space $S$) by varying hyperparameters such as maximum depth (3, 5, 7, 9 and 11), learning rate (0.01, 0.05 and 0.1), and the number of estimators  (100, 250, 500, 750 and 1000). 
 For decision trees, we used fewer configurations. 
 We partitioned the data using a 70/30 train-test partition, and recorded the performance metrics at the instance level on the test set. 
Therefore, each row $\langle \vect{x},\vect{s},\hat{y}, y\rangle$ of the test set consists of a task instance representation \vect{x} and a model configuration $\vect{s}$, with the corresponding predicted and actual results. In the case of classification, each row is defined as $\langle\vect{x}, \vect{s}, \vect{r}, y\rangle$, with $\vect{x}$ and $\vect{s}$ having the same definition, $\vect{r}$ being the predicted vector or probabilites and $y$ being the observed class. 
These results form the dataset $D$ that will serve as the training (and testing) dataset for the assessors. 

\paragraph{Splitting strategy} To prevent contamination and ensure the validity of our evaluation, we split $D$ into assessor training and testing sets (70/30) based on the instance identifiers $x_{\mathrm{id}}$. Instances with the same $x_{\mathrm{id}}$ are not shared between the assessor's training set $D_{\text{train}}$ and testing set $D_{\text{test}}$, even across different model configurations, so we ensure that assessors do not see any instance during training that they will be tested on. 

\paragraph{Assessor models and training}  The training process for the assessors is defined as follows: 
given a pair of target and proxy losses ($L_{\target}$ and $L_{\proxy}$, respectively), we train two assessors independently:

\begin{enumerate}
    \item The \textit{target assessor}: this assessor is trained to directly predict the target loss, using the tuple $\langle \vect{x},\vect{s}, L_{\target}(\hat{y}, y)\rangle$. No output post-processing is required.
    \item The \textit{proxy assessor}: this assessor is trained to predict the proxy loss $L_{\proxy}$, using the tuple $\langle \vect{x},\vect{s}, L_{\proxy}(\hat{y}, y)\rangle$. 
    The output is then transformed into the target loss, via the corresponding transformation function $f$.

\end{enumerate}

We employ various models as assessors, including \model{XGBoost}~\cite{XGB}, \model{linear regression}~\cite{lr}, \model{feed-forward neural networks}~\cite{nn}, and \model{Bayesian ridge regression}~\cite{bayreg}, to account for the different strategies these models use to solve tasks \cite{fabra2020family,fabra2024unveiling}, testing whether our results hold independently of the choice of assessor model. For the classification tasks, scores are still numerical and continuous, so we also use \model{XGBoost Regressor} as an assessor. These assessor models are trained independently of the base models used to generate the losses.


\paragraph{Evaluation metrics} 
We evaluate the relationship between the target and proxy assessors by calculating the Spearman's correlation coefficient $\rho$. To assess the statistical significance of 
the differences in $\rho$,s we establish 
95\% confidence intervals using a bootstrapping approach \cite{bootstrap}. We consider the 
differences between the proxy and target assessors statistically significant when these confidence intervals do not overlap. Furthermore, we quantify the performance of the proxy assessor relative to the target assessor by counting the number of datasets (out of the 10 in total for regression and classification, respectively) in which the proxy assessor achieves higher $\rho$ values. When the differences are not statistically significant, as indicated by overlapping confidence intervals, we categorise these cases as ties. This counting is formulated as the following score: $\#wins + \#ties + \#losses$, so that every win grants 1 point, every tie 0 points and every loss $-1$ points. 
Our score range goes from $-10$ (if the proxy assessor loses all 10 records) to 10 (if the proxy assessor wins all 10 records). A final aggregated score between $-1$ and 1 can be computed by obtaining the mean of these scores to assess the different approaches accounting for all datasets and all assessor model choices.


\section{Results}
\label{sect:results}

\begin{figure}[t]
    \centering

    \includegraphics[width=0.49\columnwidth]{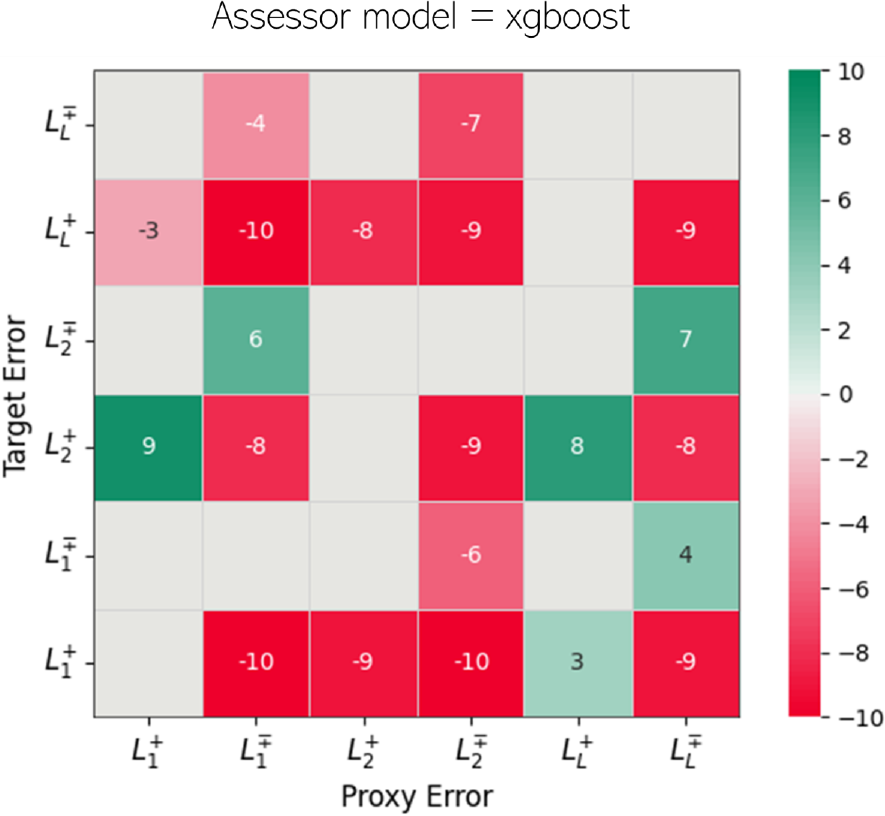}
    \includegraphics[width=0.49\columnwidth]{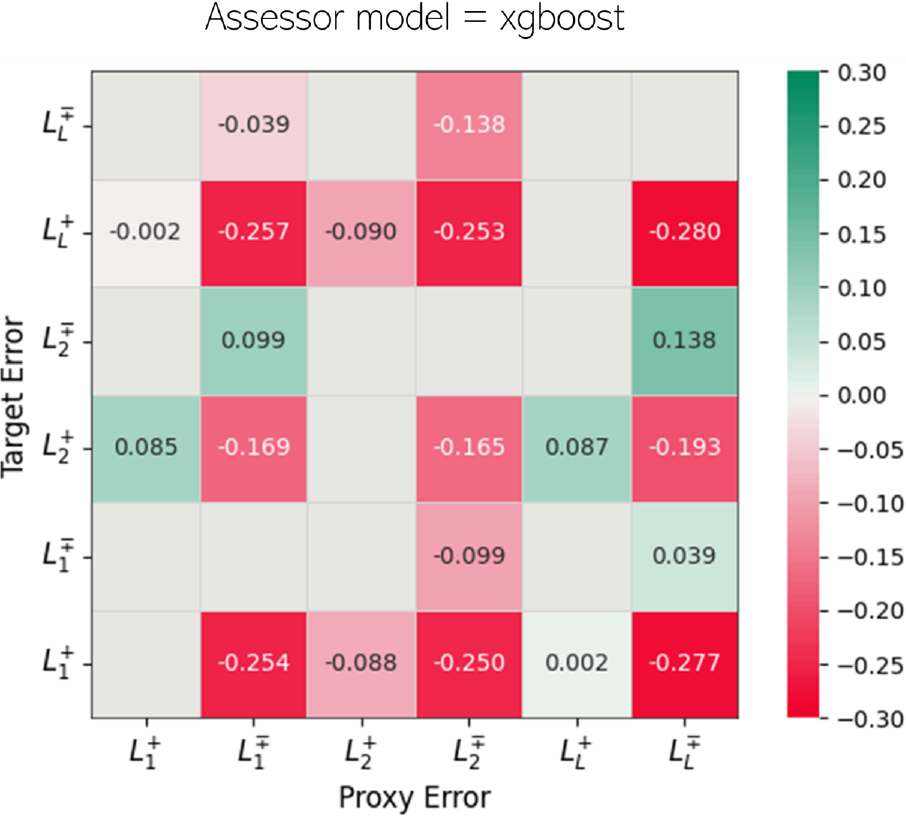}
    \caption{(Left) Score matrix for \modelc{XGBoost} assessor model. (Right) Aggregated Spearman margin matrix for \modelc{XGBoost} assessor model. In both matrices, rows represent target errors and columns proxy errors. Red values indicate poor performance from trying to predict $L_{\target}$ by learning $L_{\proxy}$. Green values show instances where learning from $L_{\proxy}$ is better than from learning directly from $L_{\target}$}
    \label{fig:xgb_score_matrix}
\end{figure}

\subsection{Regression}

Figure~\ref{fig:xgb_score_matrix} (left) shows the scores for all datasets when the assessor model of choice is \model{XGBoost} (other assessors in Appendix~\ref{app:more_matrices} show similar results). Some interesting patterns can be seen: mainly, that learning from unsigned losses (\LsgnN, \LsgnS{} and \LsgnL) to predict their unsigned counterparts yields worse assessors than learning from \LabsN, \LabsS and \LabsL directly: for instance, when the proxy error is \LsgnS and the target error is \LabsS, the final score is $-9$ (e.g., from the 10 datasets, there is one tie -- no significant differences in Spearman correlation -- and 9 losses). This contrast is specially sharp with the simple signed error, where, in all 10 datasets, its absolute counterpart yields better results in terms of Spearman correlation $\rho$. Overall, the most underperforming proxy error is by far the signed squared error, managing scores between $-10$ and $-9$ (that means no wins at all), underperforming even when comparing it to other signed losses, indicating that it is not a good proxy loss to use in general.

\begin{figure}[t]
    \centering    
    \includegraphics[width=0.49\columnwidth]{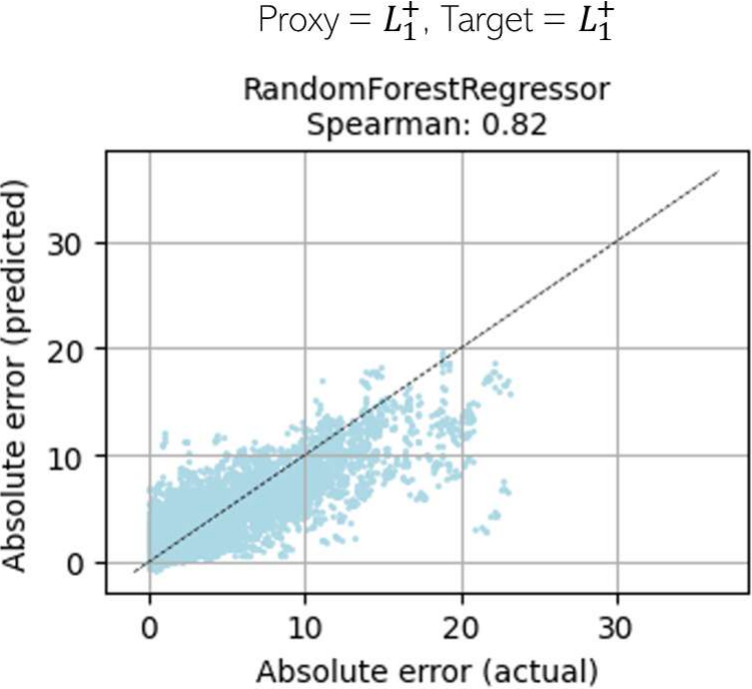}
    \includegraphics[width=0.49\columnwidth]{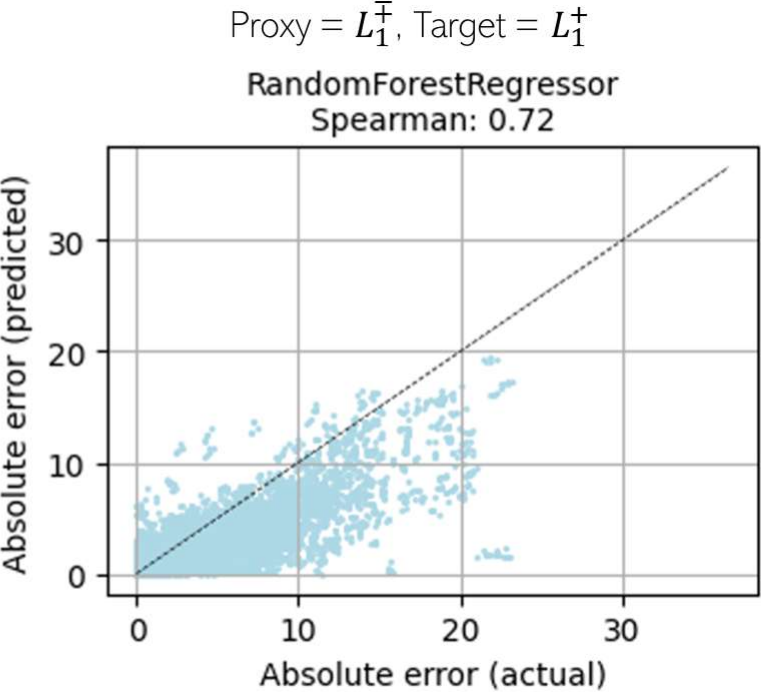}
    \caption{Scatter plots for the assessor of the Parkinson’s Disease Rating Scale for \modelc{RandomForest Regressor} base models and assessor model \modelc{XGBoost}. Because the predictions of the assessor tend to the mean, the case where the proxy is signed takes predictions towards 0, and the predictions usually fall under the diagonal}
    \label{fig:underprediction}
\end{figure}

One possible explanation for this under-performance is depicted in Figure~\ref{fig:underprediction}: assessors with signed proxies (right plot) tend to  make predictions closer to 0 (the mean), and the predictions (after the transformation $f$) underestimate the loss, even more so than those with unsigned proxies (left plot). This underestimation occurs for all the base models. For more details, see in Figure~\ref{fig:BigScatter} in Appendix~\ref{app:underpredictions}.

\begin{figure}[t]
    \centering
    \includegraphics[width=0.49\columnwidth]{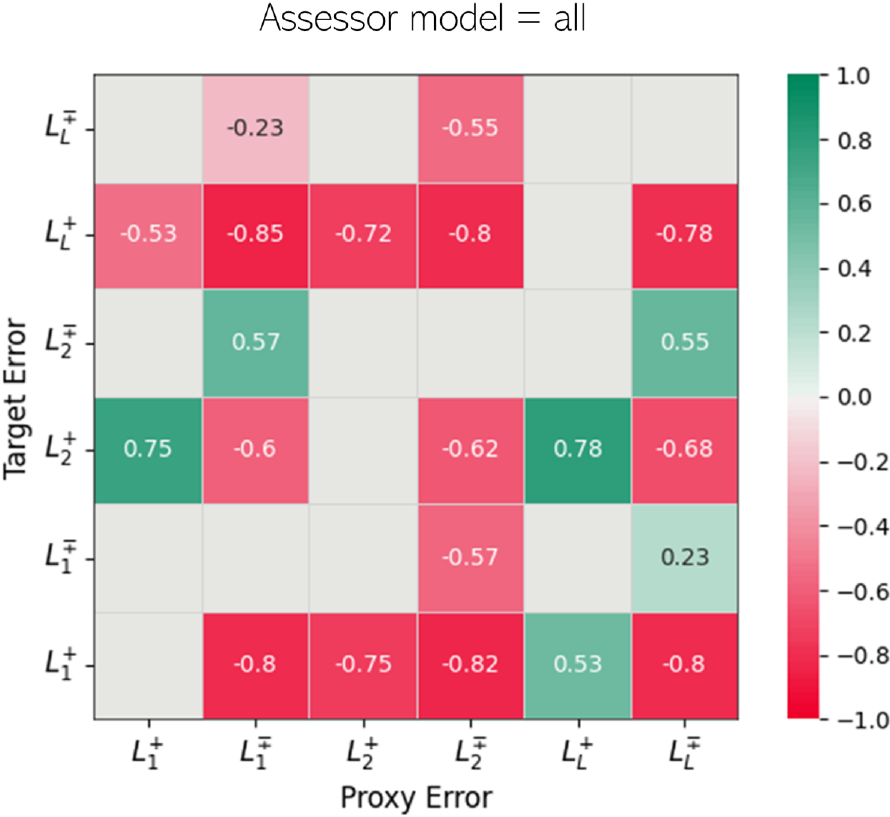}
    \includegraphics[width=0.49\columnwidth]{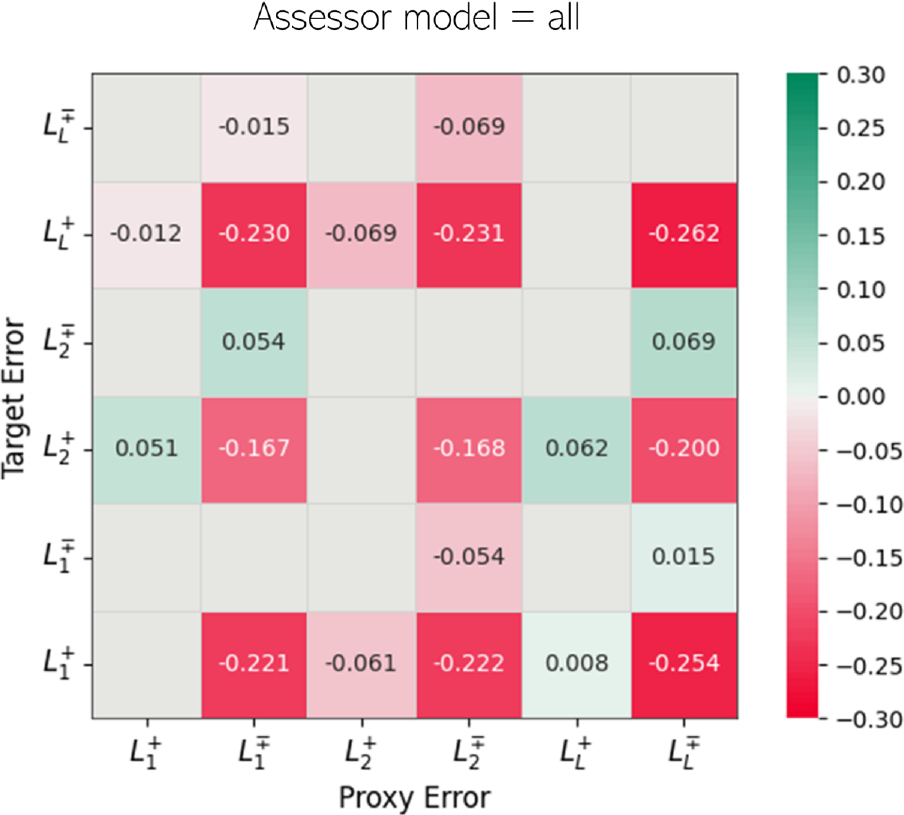}
    \caption{(Left) Mean score matrix of every possible approach between target and proxy errors. (Right) Aggregated Spearman margin matrix. In both matrices, rows represent target errors and columns proxy errors. Red values indicate poor performance from trying to predict $L_{\target}$ by learning $L_{\proxy}$. Inversely, green values show instances where learning from $L_{\proxy}$ is better than from learning directly from $L_{\target}$}
    \label{fig:mainresults}
\end{figure}

In contrast, the logistic loss shows promising results: regarding \LsgnL, when used as a proxy error to predict \LsgnN{} or \LsgnS, it outperforms the target errors (4 and 7 points, respectively). A similar pattern can be seen with \LabsL, which obtains 3 and 8 points when used as a proxy error to predict \LabsN{} and \LabsS, respectively. The simple unsigned error shows varying behaviour, outperforming \LabsS{} but not being a good proxy to predict \LabsL. 

These scores evaluate the performance of the approaches by counting the records where using a proxy loss is better than using the target loss directly. However, they are not able to quantify the magnitude of said improvement. Figure~\ref{fig:xgb_score_matrix} (right) addresses this, showing the mean Spearman difference of the 10 datasets for each combination of proxy and target error. For the computation of this mean, instances where $\rho$ is not significant are treated as having a difference equal to 0.

We see a similar behaviour to that depicted in the score matrix, although with some appreciations, specially regarding the logistic errors, where the differences are not as big as the scores matrix may suggest. The signed logistic loss presents the highest differences of the signed errors, although it manages to be a better proxy than the unsigned squared error.

These patterns are independent of the model chosen as assessor, as seen in Figure~\ref{fig:mainresults}, where a mean score taking into account all datasets and assessor models is computed, resulting in values between $-1$ and $1$, with similar interpretation as when only analysing one assessor model. Equally, Spearman differences are computed for all datasets and assessor models, with similar patterns emerging in both matrices as the ones in Figure~\ref{fig:xgb_score_matrix}. See Appendix~\ref{app:more_matrices} to see score matrices of other assessor models.

\begin{figure}[t]
    \centering
    \includegraphics[width=1\linewidth]{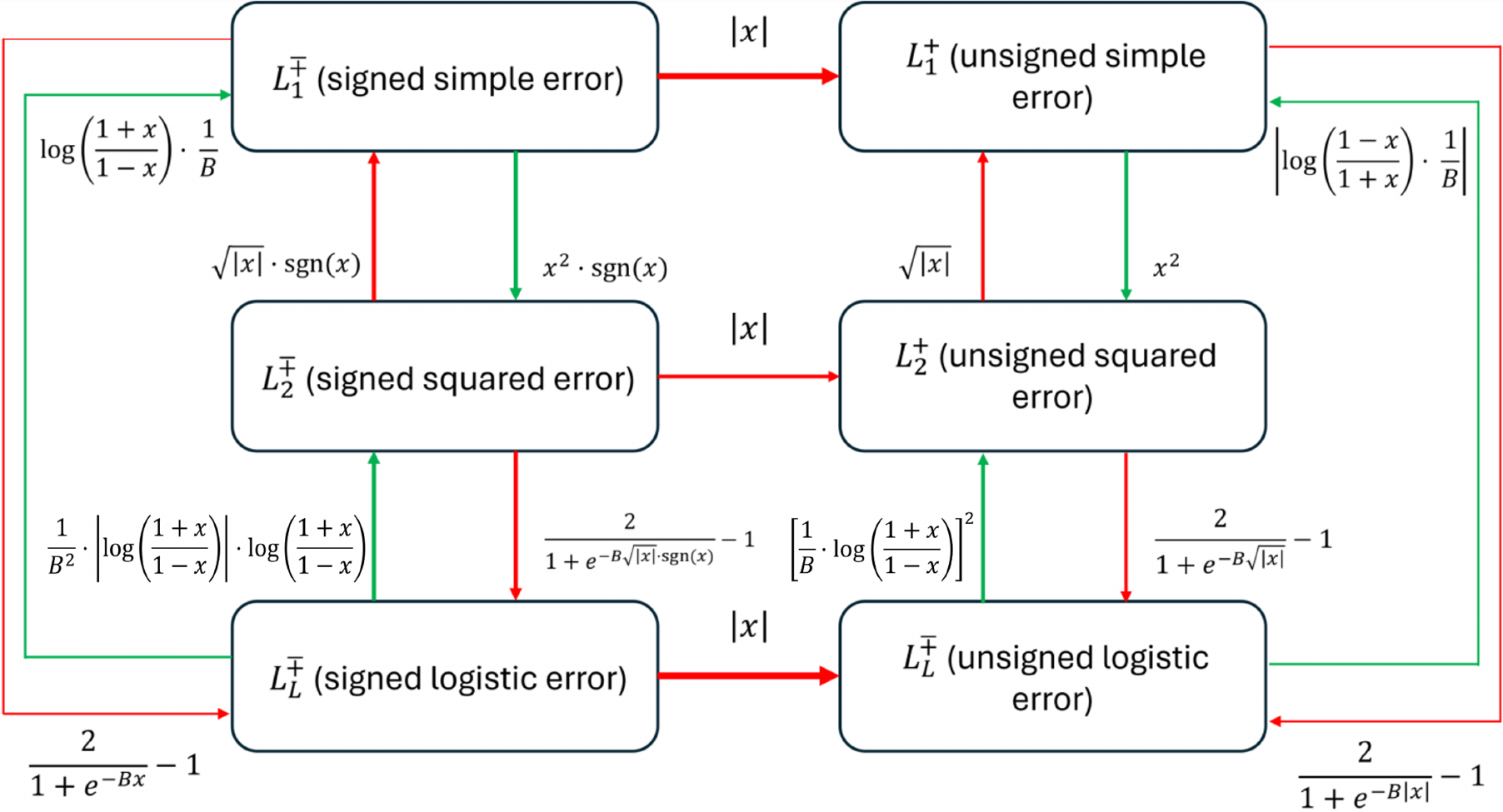}
    \caption{Which assessor metric to optimise? Signed and absolute versions (same metric) are arranged horizontally (the mapping is non-monotonic, so only one direction is possible); different metrics with monotonic transformations are aligned vertically. Arrows, color-coded in green (better) or red (worse), indicate the relationship from proxy to target metrics, with width representing the Spearman correlation margin. 
    Diagonal transformations (e.g., from signed simple error to unsigned squared error) are omitted for clarity, but displayed 
    in Figure~\ref{fig:mainresults}.}
    \label{fig:mainfigure}
\end{figure}

Figure~\ref{fig:mainfigure} summarises the results of this paper for regression by comparing most of the pairs between target and proxy losses (shown in Spearman correlation margin). We can now see more clearly that the logistic loss wins over all the other losses in its column. There also appears to be some sense of transitivity between errors: for instance, training an assessor with the signed squared error as the proxy loss to predict the target loss unsigned simple error, there is a path (two paths, in fact), that say this proxy assessor would be worse than training directly with the target loss. As shown in Figure~\ref{fig:mainresults}, this is correct. This property holds for all pairs of losses in the diagram. In cases where the arrows conforming a path are of different colours, the `strength' of the arrows (differences in $\rho$, as shown in Figure~\ref{fig:mainresults}) would dictate the final performance of the assessor.

\subsection{Classification}

\begin{figure}[t]
    \centering
    \includegraphics[width=1\linewidth]{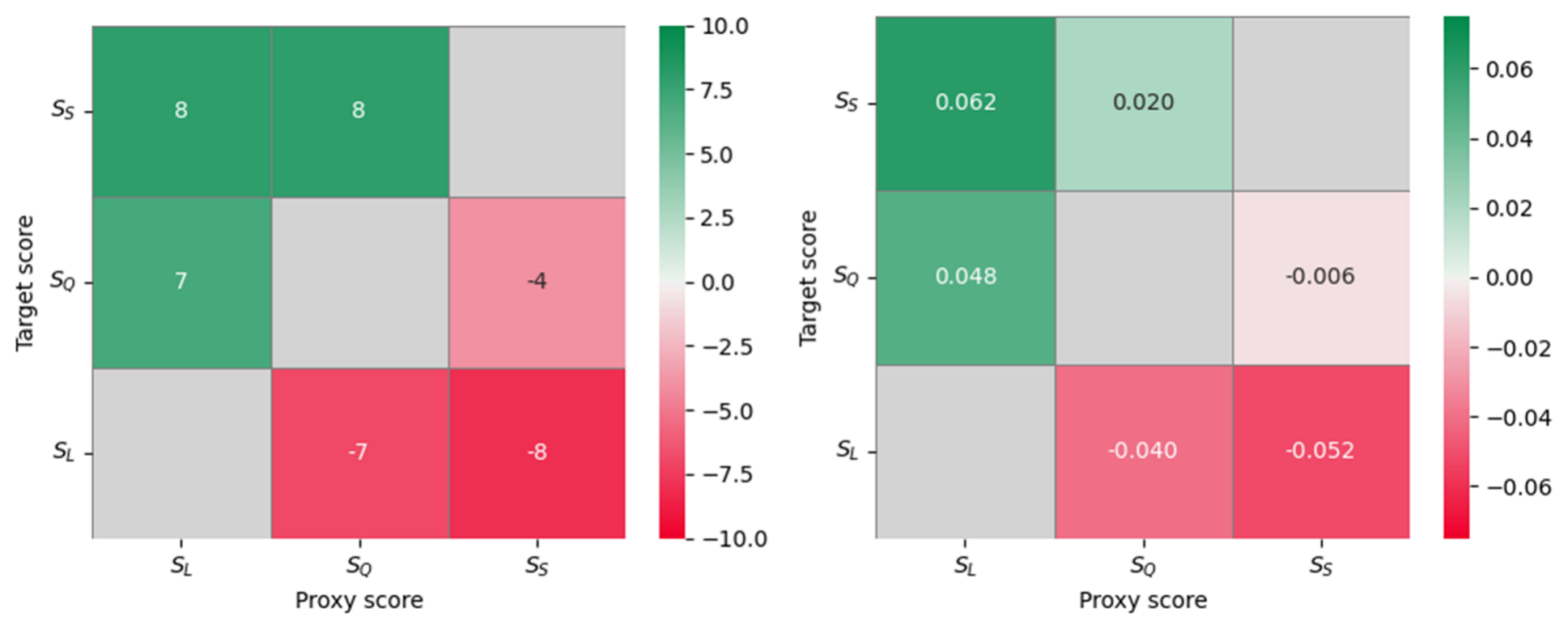}
    \caption{(Left) Score matrix for \modelc{XGBoost} assessor model. (Right) Aggregated Spearman margin matrix for \modelc{XGBoost} assessor model. In both matrices, rows represent target scores and columns proxy scores. Red values indicate poor performance from trying to predict $L_{\target}$ by learning $L_{\proxy}$. Green values show instances where learning from $L_{\proxy}$ is better than from learning directly from $L_{\target}$}
    \label{fig:classscores}
\end{figure}

Figure~\ref{fig:classscores} (left) shows the scores aggregated over all datasets and assessor models. Similar to regression, we see some scores that are good proxies: when using the logarithmic score $S_L$ as the proxy, the assessor often outperforms the target assessor for both the quadratic score $S_Q$ (8 out 10 datasets) and the spherical score $S_S$ (9 out of 10 datasets). 
Using the quadratic score $S_Q$ as a proxy to predict $S_S$ also resulted in improved performance, with positive scores in 7 out of 10 datasets. 
The spherical score $S_S$ was generally a poor proxy for predicting $S_L$ or $S_Q$.

\begin{figure}[t]
    \centering
    \includegraphics[width=1\linewidth]{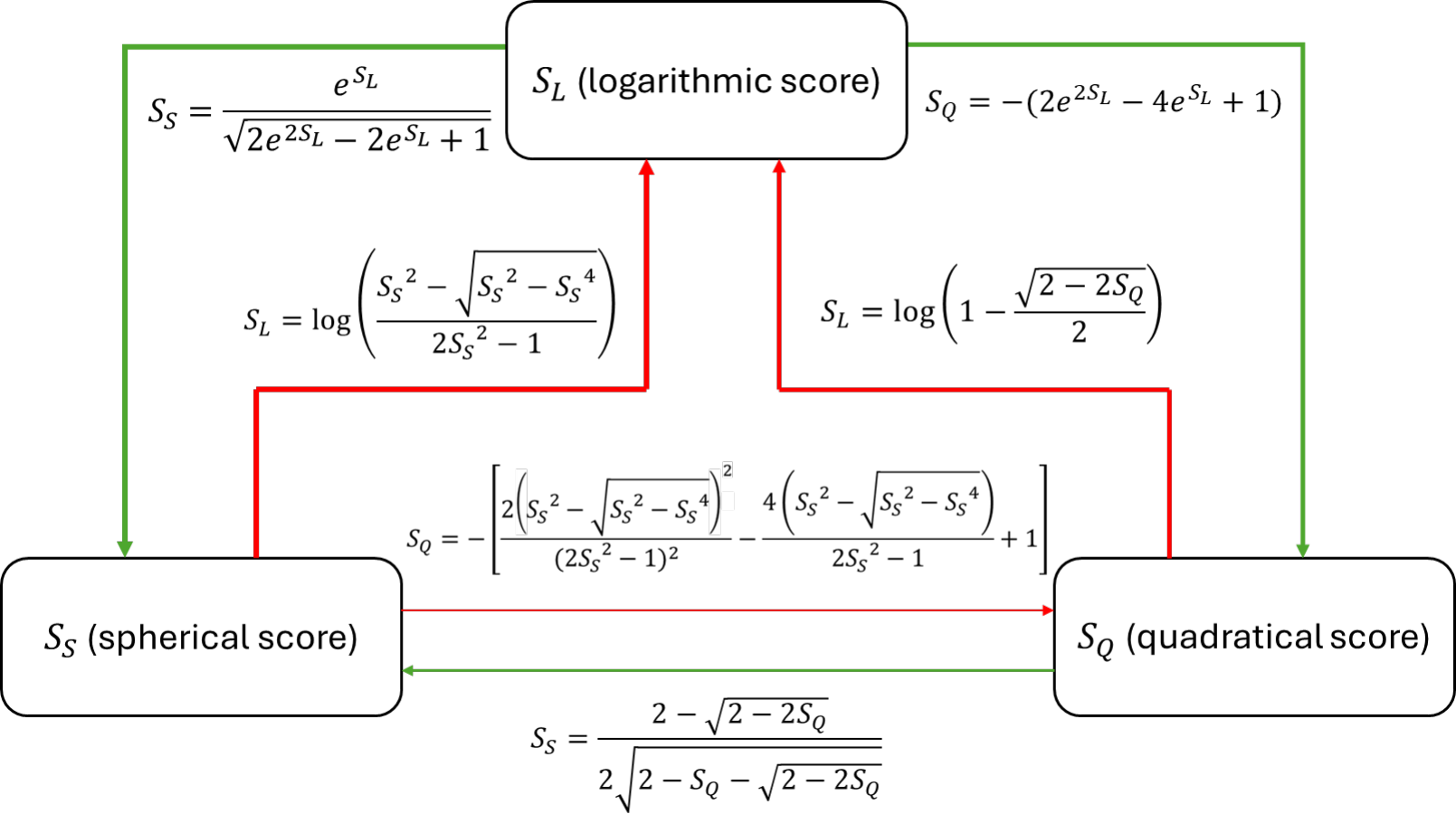}
    \caption{Similar to Figure~\ref{fig:mainfigure}, arrows indicate proxy to target score: green for better, red for worse when optimising target score; arrow width indicates Spearman correlation margin. 
    }
    \label{fig:mainclassificationdiagram}
\end{figure}

Figure~\ref{fig:classscores} (right) shows the magnitude of these 
improvements in Spearman $\rho$, establishing the logarithmic score as the best score to use as a proxy. Looking at the differences also nuances the disadvantage of using $S_S$ as a proxy to predict $S_Q$,  since although the score is $-4$, the Spearman $\rho$ difference is $-0.006$. 
Finally, Figure~\ref{fig:mainclassificationdiagram} 
illustrates the relationships between scores and their effectiveness 
as proxy metrics. It reveals patterns similar to 
those observed 
in regression, 
particularly highlighting the transitive nature of the scores.

\section{Discussion}

AI assessors represent a second-order estimation problem whose goal is to predict a loss or utility function, for any new example and base subject model. This is much more flexible than uncertainty self-estimation because we can choose the metric of the assessor to be different from the ones the base models are optimised for or evaluated. Still, in this context it may seem natural to build an assessor to optimise for the target loss. However, we see that some other proxy losses may be more effective. In regression, looking at the distribution of residuals (Figure \ref{fig:base} for a selection, Figure \ref{fig:histRegr} in the Appendix for all), one explanation may be found in a double penalisation of high residuals (e.g., for outliers). That indicates that for convex loss functions used at the first-order level (base models) we may benefit for concave loss functions at the second level that compensate for the weight in the extremes of the distribution. For classification, the phenomenon could be the reverse (Figure \ref{fig:base_r_target} for a selection, Figure  \ref{fig:histClass} in the Appendix for all) we have losses for the model in the interval $[0,1]$ not really penalising enough, and the logarithmic transformation of the principals can give more relevance to those errors.

In this paper, we chose regression problems for this first analysis of proxy losses for assessors because loss functions for regression are well known, generally continuous, and the most common one, the squared error, augments the weight of the extremes. Having explored this question for binary classification and having obtained similar results, this suggests that similar exploration for multiclass classification, and especially for losses in structured or generative tasks, 
could be done following the methodology in this paper. Similarly, in situations where a metric is composed of several parts, e.g., components in a toxicity metric or precision and recall in the F1 score, it may make more sense to estimate the components (or some monotonic transformations of the components) with separate assessors and then integrate the prediction of the overall metric. Overall, this paper opens a wide range of options for exploring the impact of loss and utility metrics when building assessors. 

\section*{Acknowledgements}

This work was funded by valgrAI, the Future of Life Institute, FLI, under grant RFP2-152, the EU (FEDER) and Spanish grant RTI2018-094403-B-C32 funded by  MCIN/AEI/10.13039/501100011033 and by CIPROM/2022/6 (FASSLOW) funded by Generalitat Valenciana, and Spanish grant PID2021-122830OB-C42 (SFERA) funded by MCIN/AEI/10.13039/501100011033 and "ERDF A way of making Europe"


\bibliography{references}
\bibliographystyle{icml2025}

\newpage
\appendix
\onecolumn
\section{Detailed Exploration of Scoring Rules and Loss Functions and Loss Functions}\label{app:lossscoreexploration}

In this appendix, we provide a comprehensive analysis of the scoring rules used for classification and the loss functions used for regression in our study. We include graphical representations of these functions and derive their mathematical relationships. These derivations demonstrate monotonic transformations between different loss and scoring functions, allowing predictions to be converted from one function to another. This capability is particularly useful for training assessors, as it allows us to train a assessor on one loss or score function and then transform their output to estimate a different function. 

\subsection{Proper Scoring Rules for Classification}

\begin{figure}[H]
    \centering
    \includegraphics[width=0.85\linewidth]{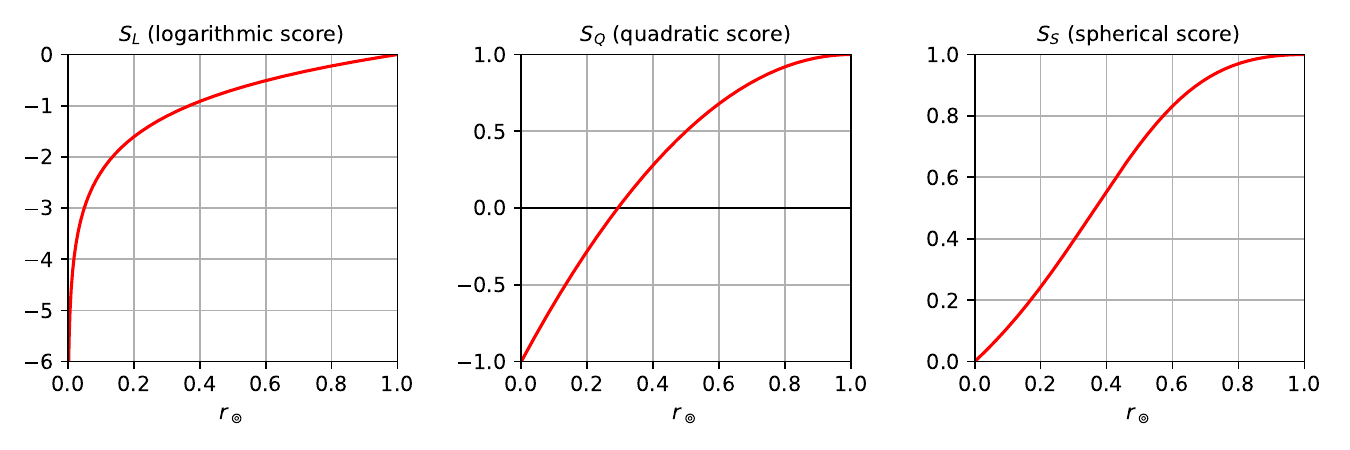}
    \caption{Functional representation of the three scoring rules we use in this paper, logarithmic ($S_L$), quadratic ($S_Q$) and spherical ($S_S$) scores.}
    \label{fig:scoringrules}
\end{figure}

In probabilistic classification, we evaluate the quality of predicted probability distributions over classes. Proper scoring rules are functions that encourage truthful reporting of probabilities. We consider  the following proper scoring rules: Logarithmic Score ($S_L$), Quadratic (Brier) Score ($S_Q$) and Spherical Score ($S_S$). All three scoring rules are proper, meaning they are minimised when the predicted probabilities match the true probabilities. 
Figure~\ref{fig:scoringrules} illustrates the behaviour of the three scoring rules as a function of $r_{\circledcirc}$, the predicted probability of the correct class.

\subsubsection{Derivation of monotonic transformations between scores}

We derive mathematical relationships between the scoring rules to understand how they are related and to facilitate transformations from one scoring rule to another. Let's start remembering the three proper scoring rules:

\begin{equation*}
    S_L(\vect{r}) = ln({r_\circledcirc})
\end{equation*}

\begin{equation*}
    S_Q(\vect{r}) = 2{r_\circledcirc} - \vect{r} \cdot \vect{r}
\end{equation*}

\begin{equation*}
    S_S(\vect{r}) = \frac{{r_\circledcirc}}{||\vect{r}||}
\end{equation*}

In binary classification,  the predicted probabilities sum to 1, so we can express $\vect{r}$ in terms of $r_\circledcirc$ as $\vect{r} = (r_\circledcirc, 1 - r_\circledcirc)$ or $\vect{r} = (1- r_\circledcirc, r_\circledcirc)$, depending on which one is the correct class. However, for both cases, the product is the same ${r_\circledcirc}^2 + (1-{r_\circledcirc})^2$, and so is the norm. 

First, we expand the quadratic score:

\begin{equation}
\label{eq:15}
    S_Q(\vect{r}) = 2{r_\circledcirc} - \vect{r} \cdot \vect{r} = 2{r_\circledcirc} - [{r_\circledcirc}^2 + (1-{r_\circledcirc})^2] = -2{r_\circledcirc}^2 + 4{r_\circledcirc} - 1 =  -(2{r_\circledcirc}^2 - 4{r_\circledcirc}+1)
\end{equation}

The same can be done with the spherical score:

\begin{equation}
\label{eq:16}
        S_S(\vect{r}) = \frac{{r_\circledcirc}}{||\vect{r}||} = \frac{{r_\circledcirc}}{\sqrt{{r_\circledcirc}^2 + (1-{r_\circledcirc})^2}} = \frac{{r_\circledcirc}}{\sqrt{2{r_\circledcirc}^2 - 2{r_\circledcirc} + 1}}
\end{equation}

From now on we will denote the scoring rules without their input ${r_\circledcirc}$ for clarity (that is, for instance, $S_L$ instead of $S_L({r_\circledcirc})$). We can derive the transformations between scoring rules as follows, starting with the logarithmic score:

\begin{equation*}
    S_L = \ln({r_\circledcirc}) \rightarrow {r_\circledcirc} = e^{S_L}
\end{equation*}

Substituting in Eq.~\ref{eq:15} and Eq.~\ref{eq:16}, respectively:

\begin{equation}
\label{eq:17}
    S_Q = -(2e^{2S_L} - 4e^{S_L} + 1)
\end{equation}

\begin{equation}
\label{eq:18}
    S_S = \frac{e^{S_L}}{\sqrt{2e^{2{S_L}} - 2e^{S_L} + 1}}
\end{equation}

Obtaining the quadratic and spherical scores in terms of the logarithmic score. We can do the same (obtaining $r_\circledcirc$ in terms of one score and then substituting the others), but they are not as straightforward as those with the logarithmic score. We start with the quadratic score $S_Q$:

\begin{equation}
\label{eq:19}
    S_Q = -(2{r_\circledcirc}^2 - 4{r_\circledcirc}+1) \rightarrow r_\circledcirc = 1 \pm \frac{\sqrt{2 - 2S_Q}}{2}
\end{equation}

To know which branch to take, let's recall first the domain and range of $S_Q$:

\begin{equation*}
    \text{Dom}(S_Q) = [0, 1], \quad \text{Rng}(S_Q) = [-1, 1]
\end{equation*}

We can discard the positive branch of Eq.~\ref{eq:19} since a value of $-1$ in $S_Q$ would yield a value outside the domain of $S_Q$:

\begin{equation*}
    \text{if} \quad r_\circledcirc = 1 + \frac{\sqrt{2-2S_Q}}{2}, \quad \text{if} \quad S_Q = -1 \rightarrow r_\circledcirc = 2 \notin \text{Dom}(S_Q)
\end{equation*}

In fact, only the negative branch is the inverse of $S_Q$:

\begin{equation}
\label{eq:20}
    r_\circledcirc = 1 - \frac{\sqrt{2 - 2S_Q}}{2}
\end{equation}

Using Eq.~\ref{eq:20}, we can now substitute in $S_L$ and $S_S$, respectively:

\begin{equation}
\label{eq:21}
    S_L = \ln \left(1 - \frac{\sqrt{2 - 2S_Q}}{2}\right)
\end{equation}

\begin{equation}
\label{eq:22}
    S_S = \frac{1 - \frac{\sqrt{2 - 2S_Q}}{2}}{2\left(1 - \frac{\sqrt{2 - 2S_Q}}{2}\right)^2 - 2\left(1 - \frac{\sqrt{2 - 2S_Q}}{2}\right) + 1} = \frac{1 - \frac{\sqrt{2 - 2S_Q}}{2}}{\sqrt{2-S_Q - \sqrt{2-2S_Q}}} = \frac{2-\sqrt{2-2S_Q}}{2\sqrt{2-S_Q-\sqrt{2-2S_Q}}}
\end{equation}

Obtaining the logarithmic and spherical scores in terms of the quadratic score.

Finally, we can do the same for the spherical score:

\begin{equation}
\label{eq:23}
    S_S = \frac{{r_\circledcirc}}{\sqrt{2{r_\circledcirc}^2 - 2{r_\circledcirc} + 1}} \rightarrow r_\circledcirc = \frac{{S_S}^2 \pm \sqrt{{S_S}^2-{S_S}^4}}{2{S_S}^2-1}
\end{equation}

\begin{figure}[H]
    \centering
    \includegraphics[width=0.4\columnwidth]{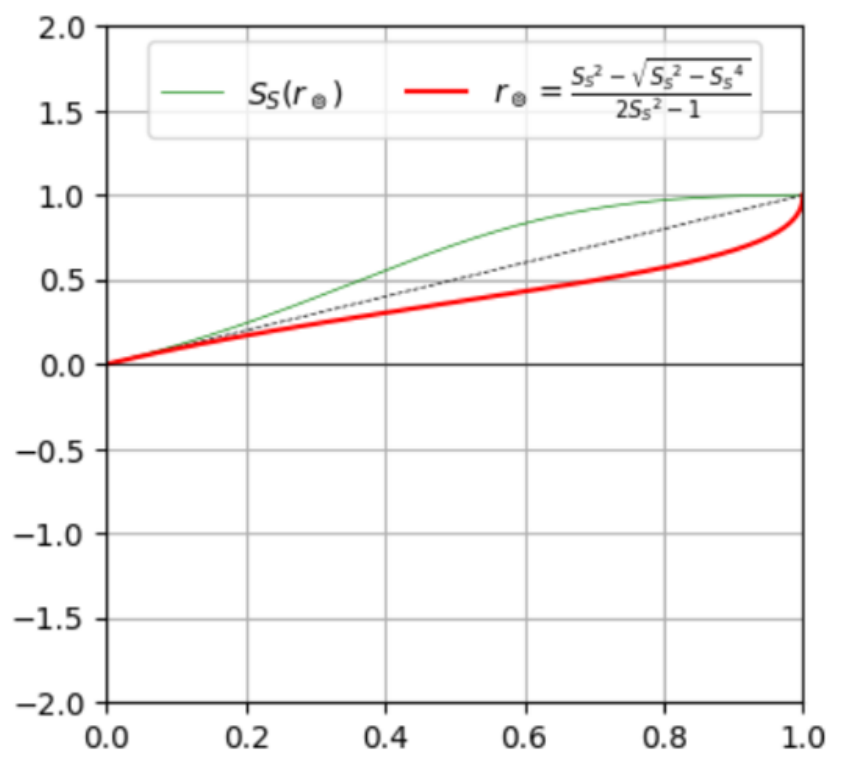}
    \includegraphics[width=0.38\columnwidth]{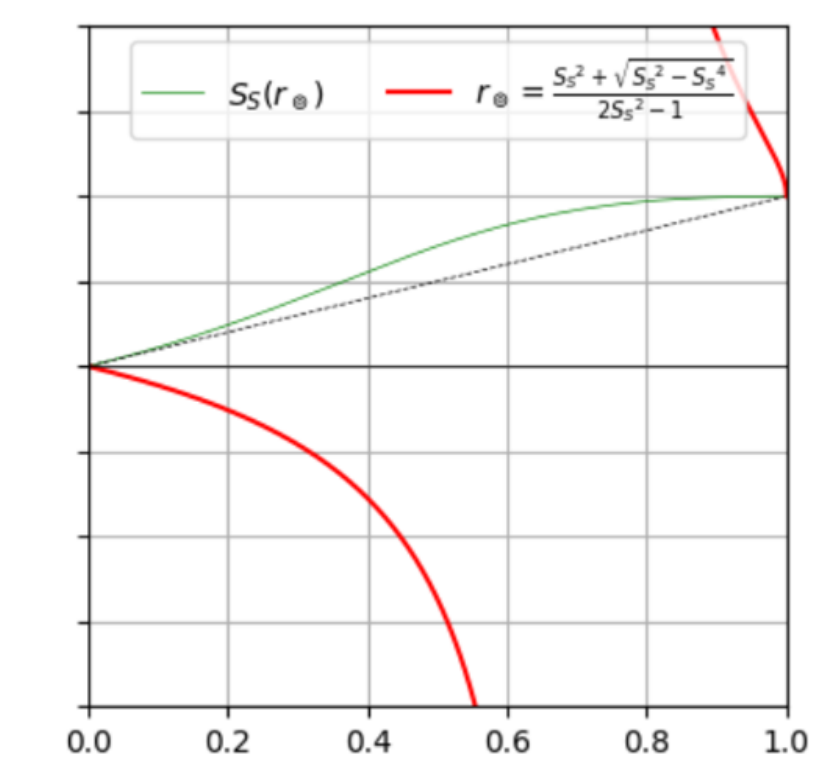}
    \caption{Graphical representation of both branches of Eq.~\ref{eq:23}. The positive branch shows values outside the range of $S_S$.}
    \label{fig:sphere}
\end{figure}

In this case, deciding which branch is the inverse of $S_S$ is a bit trickier, since substitution of the limits of the domain of $S_S$ $([0,1])$ does not yield values outside the range $([0,1])$. However, the graphical representation of both (see Figure~\ref{fig:sphere}) shows that the negative one is the inverse of $S_S$.

As such, we end up with:

\begin{equation}
    S_S = \frac{{r_\circledcirc}}{\sqrt{2{r_\circledcirc}^2 - 2{r_\circledcirc} + 1}} \rightarrow r_\circledcirc = \frac{{S_S}^2 - \sqrt{{S_S}^2-{S_S}^4}}{2{S_S}^2-1}
\end{equation}

And by substitution, in Eq.~\ref{eq:15} and Eq.~\ref{eq:16}, respectively, we obtain the remaining transformation functions:

\begin{equation}
\label{eq:25}
    S_L = \ln\left(\frac{{S_S}^2 - \sqrt{{S_S}^2-{S_S}^4}}{2{S_S}^2-1}\right)
\end{equation}

\begin{equation}
\label{eq:26}
    S_Q = -\left[\frac{2\left({S_S}^2-\sqrt{{S_S}^2-{S_S}^4}\right)^2}{\left(2{S_S}^2-1\right)^2} - \frac{4\left({S_S}^2-\sqrt{{S_S}^2-{S_S}^4}\right)}{2{S_S}^2-1} + 1\right]
\end{equation}

Equations~\ref{eq:15},~\ref{eq:16},~\ref{eq:21},~\ref{eq:22},~\ref{eq:25} and~\ref{eq:26} are shown in the main text.

\subsection{Loss Functions for Regression}

\begin{figure}[H]
    \centering
    \includegraphics[width=0.85\linewidth]{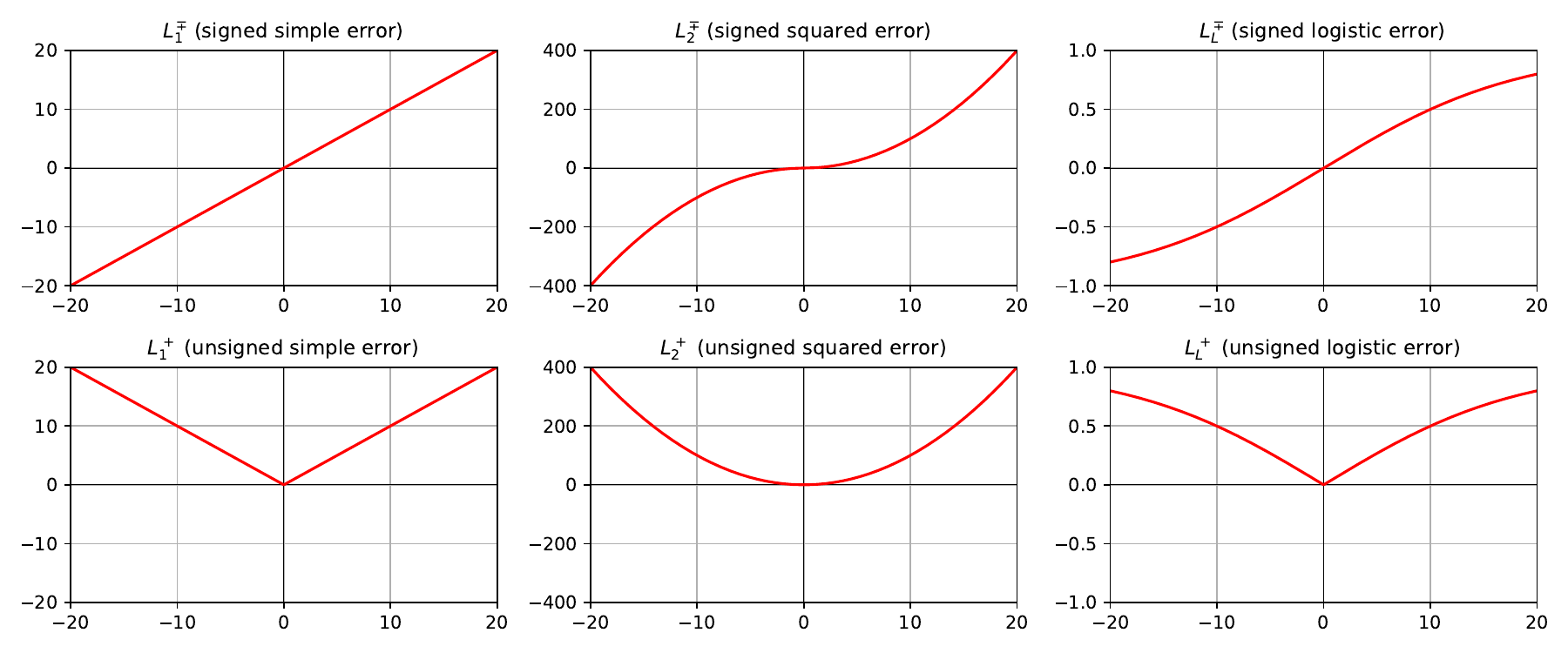}    
    \caption{Functional representation of the six losses we use in this paper, signed ($\LsgnN$, $\LsgnS$ and $\LsgnL$) and unsigned ($\LabsN$, $\LabsS$ and $\LabsL$).}
    \label{fig:losses}
\end{figure}

The performance of regression models is typically evaluated using various loss functions that measure the discrepancy between the predicted values $\hat{y}$ and the true values $y$. In our study, we consider the Signed Simple Error ($\LsgnN$), which captures both the magnitude and direction of the error; Signed Squared Error ($\LsgnS$), which amplifies the impact of larger errors; and the Signed Logistic Loss ($\LsgnL$), which  adjusts the steepness of the logistic function based on the average absolute error. All are defined as functions of the residual $e = \hat{y} - y$. The corresponding unsigned loss functions are obtained by taking the absolute value of the signed losses ($\LabsN$, $\LabsS$ and $\LabsL$):

Figure~\ref{fig:losses} illustrates the behaviour of the six loss functions as a function of the residual $e$. The signed losses ($\LsgnN$, $\LsgnS$, $\LsgnL$) capture both magnitude and direction, while the unsigned losses ($\LabsN$, $\LabsS$, $\LabsL$) capture only the magnitude.

\subsubsection{Derivation of monotonic transformations between losses}

We now derive mathematical relationships between the loss functions to understand their interrelationships and to facilitate transformations from one loss function to another. We will start with the signed losses defined in Section~\ref{sect:problem_rep} in the main text:

\begin{equation*}
\label{eq:27}
    \LsgnN(\hat{y}, y) = \hat{y} - y
\end{equation*}

\begin{equation*}
\label{eq:28}
    \LsgnS(\hat{y}, y) = (\hat{y} - y) \cdot |\hat{y} - y|
\end{equation*}

\begin{equation*}
    \LsgnL(\hat{y}, y) = \frac{2}{1+e^{-B(\hat{y} - y)}} - 1, B = \frac{\ln 3}{\text{mean}_Y|\hat{y}-y|}
\end{equation*}

Similar to the proper scoring rules, we can obtain $(\hat{y} - y)$ (also known as the residual $e$) in terms of the different losses. \LsgnN already satisfies this, so we can substitute the other losses (similarly to the previous section, we will omit the residual in the following formulae for clarity):

\begin{equation}
    \LsgnS = \LsgnN \cdot |\LsgnN|
\end{equation}

\begin{equation}
    \LsgnL = \frac{2}{1+e^{-B\LsgnN}} - 1, B = \frac{\ln 3}{\text{mean}_Y|\LsgnN|}
\end{equation}

For their unsigned counterparts, we can take their definitions as the absolute value of the signed losses:

\begin{equation*}
    \LabsN = |\LsgnN|
\end{equation*}

\begin{equation*}
    \LabsS = |\LsgnS| = |\LsgnN \cdot |\LsgnN|| = (\LsgnN)^2
\end{equation*}

\begin{equation*}
    \LabsL = |\LsgnL| = \left|\frac{2}{1+e^{-B\LsgnN}} - 1\right|, B = \frac{\ln 3}{\text{mean}_Y|\LsgnN|}
\end{equation*}

Regarding representing $\hat{y}-y$ in terms of $\LsgnS$, note that:

\begin{equation}
    \hat{y} - y = \begin{cases}
                    \sqrt{\LsgnS} & \text{if $\LsgnS >= 0$}\\
                    -\sqrt{-\LsgnS} & \text{if $\LsgnS < 0$}
                  \end{cases}
\end{equation}

Which is equivalent to using the square root and the sign function:

\begin{equation}
    \hat{y} - y = \sqrt{|\LsgnS|} \cdot \text{sgn}(\LsgnS)
\end{equation}

We can now substitute in the rest of the losses:

\begin{equation}
    \LsgnN = \sqrt{|\LsgnS|} \cdot \text{sgn}(\LsgnS)
\end{equation}

\begin{equation}
    \LsgnL = \frac{2}{1+e^{-B\sqrt{|\LsgnS|}\cdot \text{sgn}(\LsgnS)}} - 1, B = \frac{\ln 3}{\text{mean}_Y\left|\sqrt{|\LsgnS|}\cdot \text{sgn}(\LsgnS)\right|}
\end{equation}

For the unsigned losses:

\begin{equation}
    \LabsN = \left|\sqrt{|\LsgnS|} \cdot \text{sgn}(\LsgnS)\right| = \sqrt{|\LsgnS|}
\end{equation}

\begin{equation}
    \LabsS = |\LsgnS|
\end{equation}

\begin{equation}
    \LabsL = \left|\frac{2}{1+e^{-B\sqrt{|\LsgnS|}\cdot \text{sgn}(\LsgnS)}} - 1\right|, B = \frac{\ln 3}{\text{mean}_Y\left|\sqrt{|\LsgnS|}\cdot \text{sgn}(\LsgnS)\right|}
\end{equation}

Finally, for the logistic loss:

\begin{align}
    \LsgnL(\hat{y}, y) = \frac{2}{1+e^{-B(\hat{y} - y)}} & - 1 \rightarrow e^{-B(\hat{y} - y)} = \frac{2}{\LsgnL + 1} - 1 \rightarrow \hat{y} - y = -\ln\left( \frac{1 - \LsgnL}{1 + \LsgnL}
 \right) \cdot \frac{1}{B} \rightarrow \notag \\
 & \rightarrow \hat{y} - y = \ln{\left(\frac{1 - \LsgnL}{1 + \LsgnL}\right)}^{-1} \cdot \frac{1}{B} = \frac{1}{B} \cdot \ln\left(\frac{1 + \LsgnL}{1 - \LsgnL}\right)
\end{align}

Substituing in Equations~\ref{eq:27} and~\ref{eq:28}, respectively:

\begin{equation}
    \LsgnN = \frac{1}{B} \cdot \ln\left(\frac{1 + \LsgnL}{1 - \LsgnL}\right)
\end{equation}

\begin{equation*}
    \LsgnS = \frac{1}{B} \cdot \ln\left(\frac{1 + \LsgnL}{1 - \LsgnL}\right) \cdot \left| \frac{1}{B} \cdot \ln\left(\frac{1 + \LsgnL}{1 - \LsgnL}\right) \right|
\end{equation*}

Since $B$ is defined as the mean of the absolute value of $(\hat{y} - y)$, it is always positive, so:

\begin{equation}
    \LsgnS = \frac{1}{B^2} \cdot \ln\left(\frac{1 + \LsgnL}{1 - \LsgnL}\right) \cdot \left| \ln\left(\frac{1 + \LsgnL}{1 - \LsgnL}\right) \right|
\end{equation}

And for their unsigned counterparts:

\begin{equation}
    \LabsN = \left| \frac{1}{B} \cdot \ln\left(\frac{1 + \LsgnL}{1 - \LsgnL}\right) \right|
\end{equation}

\begin{equation}
    \LabsS = \left| \frac{1}{B^2} \cdot \ln\left(\frac{1 + \LsgnL}{1 - \LsgnL}\right) \cdot \left| \ln\left(\frac{1 + \LsgnL}{1 - \LsgnL}\right) \right| \right| = \frac{1}{B^2} \cdot \left[ \ln\left(\frac{1 + \LsgnL}{1 - \LsgnL}\right)\right]^2
\end{equation}

\begin{equation}
    \LabsL = \left| \LsgnL \right|
\end{equation}

\section{Assessor training methodology details}\label{app:asstraining}

This appendix complements the assessor training methodology described in Section~\ref{sect:methods} complementing the main text with comprehensive details on test result generation and assessor training. For the generation of base model results, where we use five tree-based learning algorithms and varied hyperparameters to create different model configurations. Each configuration is evaluated using a dataset split into training and test sections, yielding predicted and true outputs. An evaluation dataset is then constructed, comprising input features from the original problem and model features with specific loss functions as targets. Importantly, the splitting strategy ensures no data leakage by using instance identifiers to separate training and test sets. 

\begin{figure}[!h]
    \centering
    \includegraphics[width=0.65\linewidth]{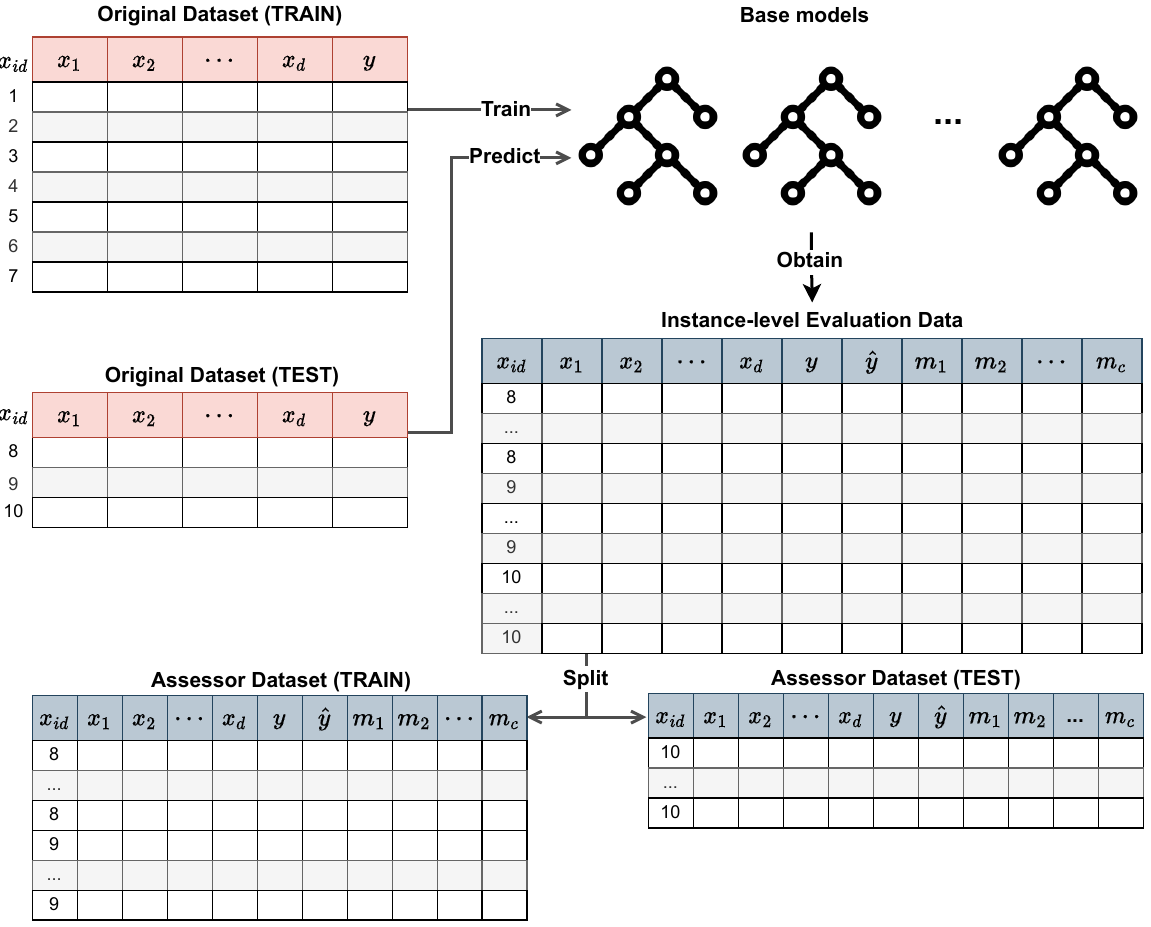}
    \caption{(Top) Procedure to obtain instance level evaluation results. In the final datasets, the original problem features $X$, as well as the model characteristics $S$, constitute an example for the assessor. 
    (Bottom) To avoid contamination, a splitting method is applied to the data, so that the assessor training does not have any $\vect{x}$ that appears in the test for the assessor, with the same or different $\vect{m}$. The instance \vect{x} identifier $x_\mathrm{id}$ is only shown for illustration, but not used in the training or evaluation of the assessor.}
    \label{fig:assessorevaldata}
\end{figure}

Figure~\ref{fig:assessorevaldata} visually shows this data generation and splitting approach, illustrating how the original problem features and model characteristics form an example for an assessor.

\newpage

\section{Residuals and Principals}\label{app:histograms}

In this appendix we plot the residuals for the regression datasets and the principals for the classification datasets used in our experiments (see Table~\ref{tab:ds}).  For each regression dataset, we plot the scatter plots of residuals (errors) $\hat{y} - y$ across all base models. For each classification dataset, we show histograms illustrating the distribution of principals ${r_\circledcirc}$ , which are the probabilities assigned by the base classifiers to the correct class.

\begin{figure}[H]
    \centering
    \resizebox{\columnwidth}{!}{%
    \addtolength{\tabcolsep}{-1.1em}
    \begin{tabular}{ccccc}
       
        \textbf{Human} & \textbf{Bank Marketing} &  \textbf{Magic gamma telescope} & \textbf{JM1} & \textbf{Nomao}\\
        \includegraphics[width=0.3\linewidth]{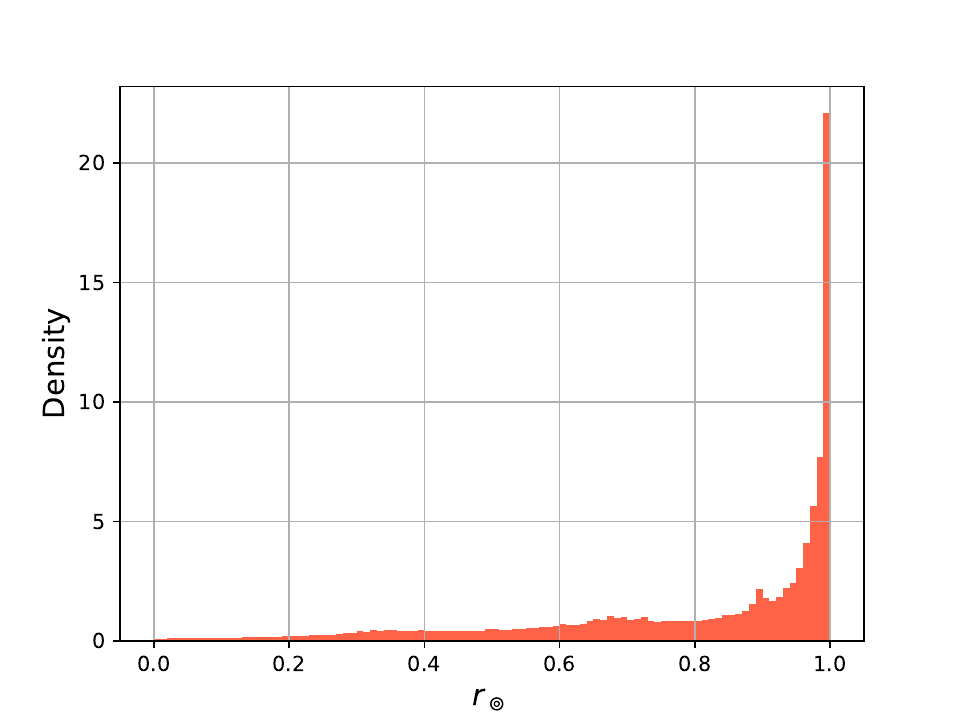} &
        \includegraphics[width=0.3\linewidth]{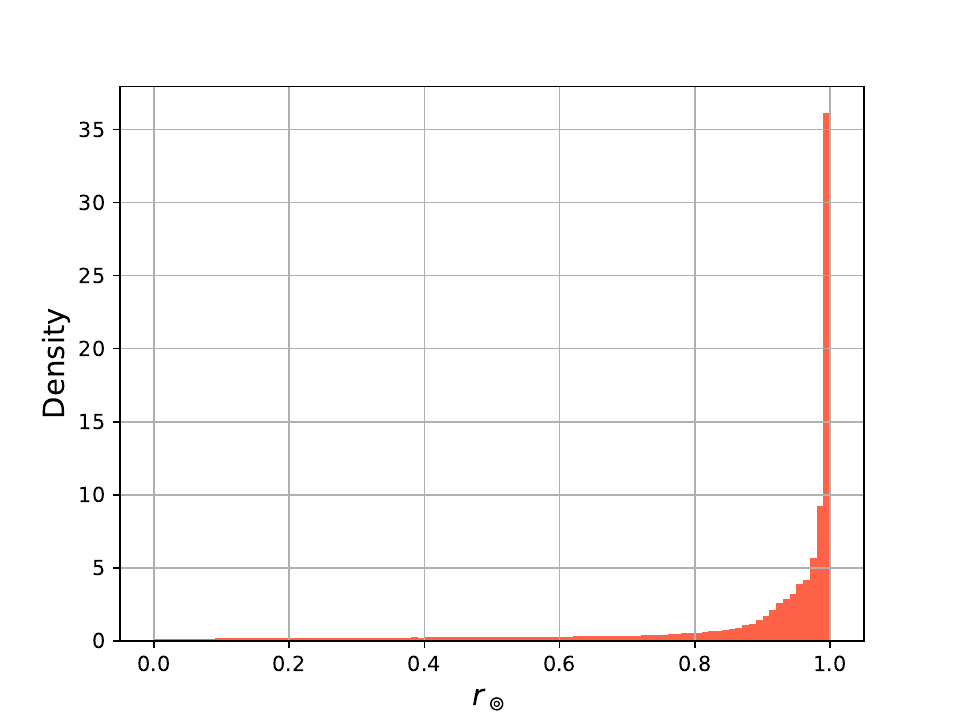} &
        \includegraphics[width=0.3\linewidth]{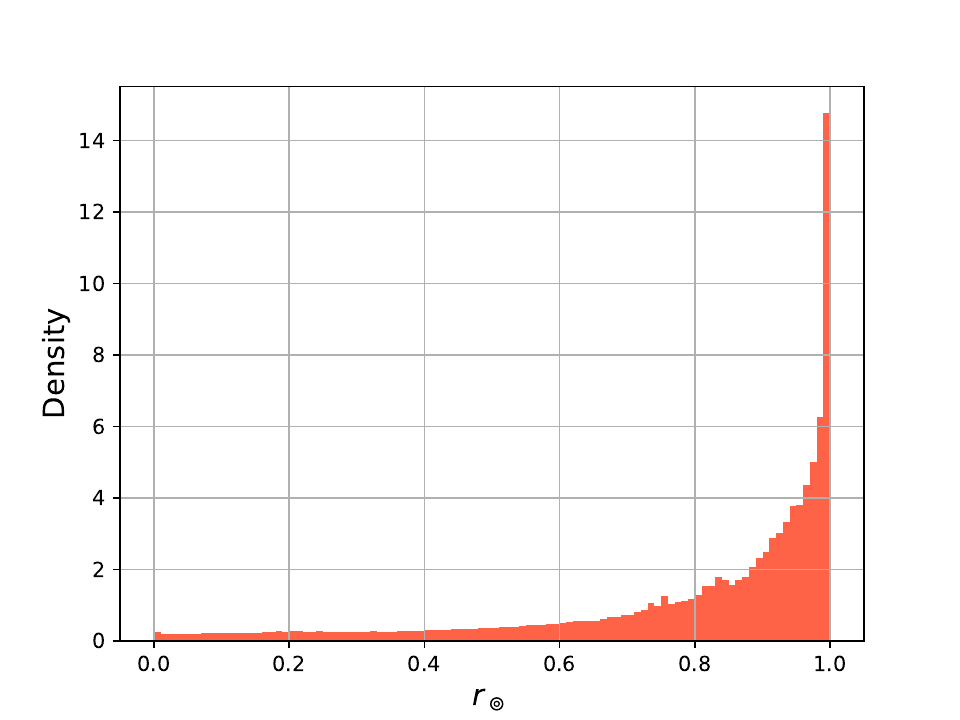} & \includegraphics[width=0.3\linewidth]{Figures/AppVI/B4_JM1.pdf} &
        \includegraphics[width=0.3\linewidth]{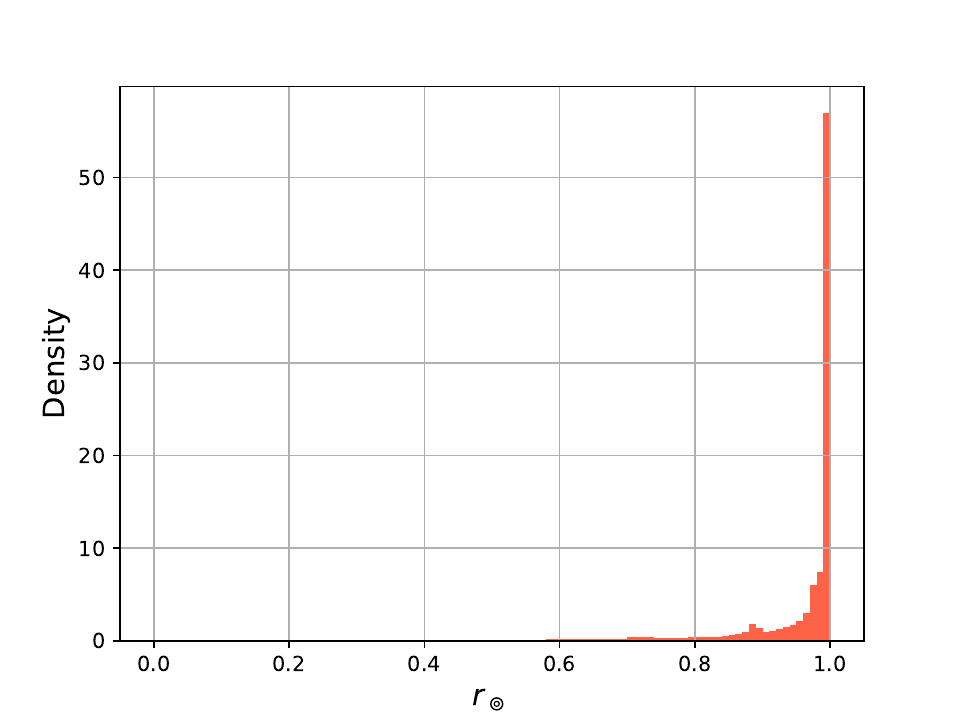}\\

        \textbf{Pen Digits} & \textbf{CDC Diabetes} & \textbf{Higgs} &  \textbf{Credit Card Fraud} & \textbf{R\&K vs K Checkmates}\\
        \includegraphics[width=0.3\linewidth]{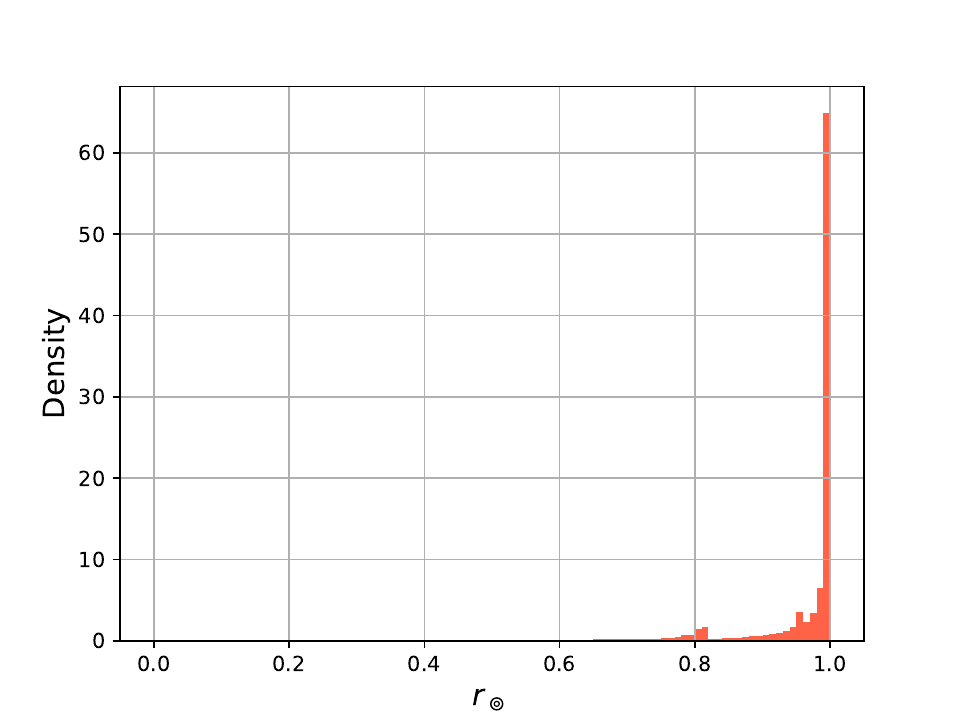} &
        \includegraphics[width=0.3\linewidth]{Figures/AppVI/B7_Diabetes.pdf} &
        \includegraphics[width=0.3\linewidth]{Figures/AppVI/B8_Higgs.pdf} &
        \includegraphics[width=0.3\linewidth]{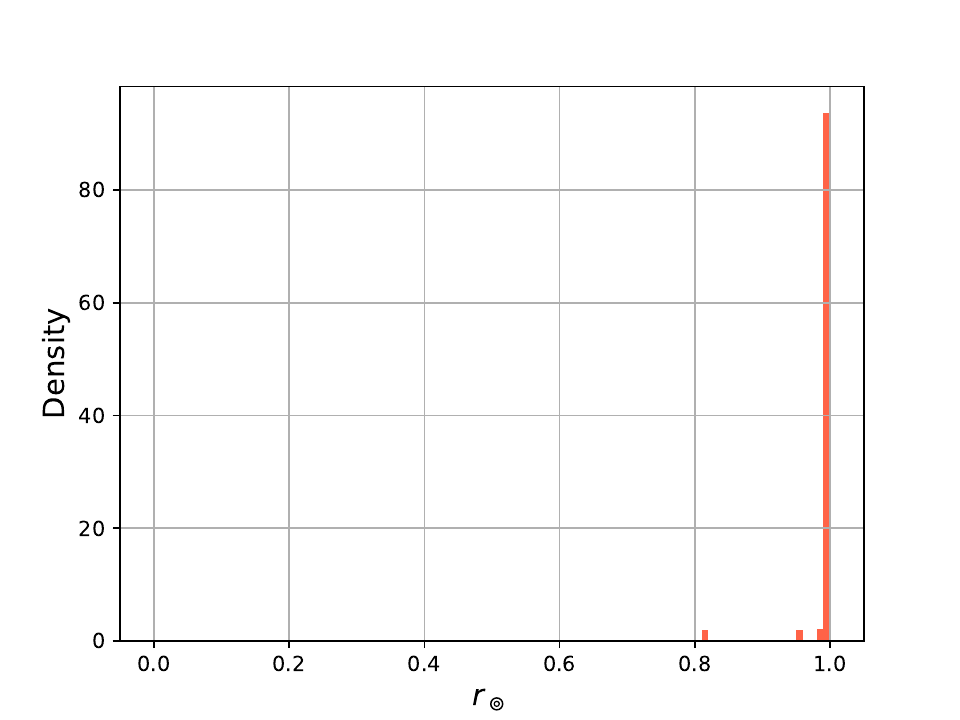} &
        \includegraphics[width=0.3\linewidth]{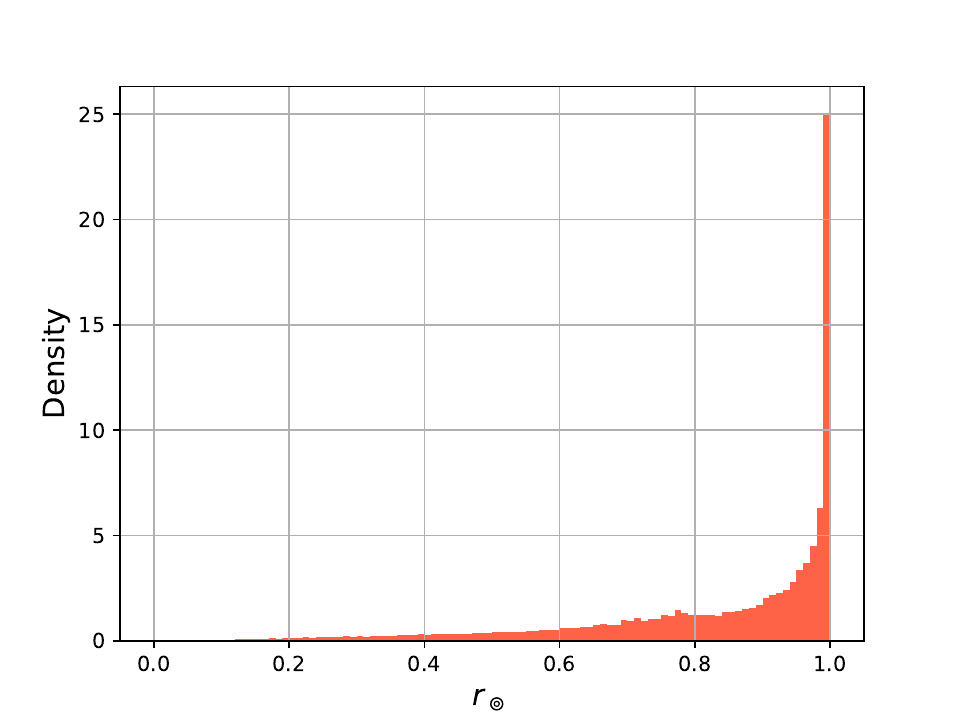}\\
    \end{tabular}
    }

    \caption{Histogram of $r_\circledcirc$. Except for \ds{Higgs}, the base models are generally good, so there are few high residuals (low $r_\circledcirc$). }
    \label{fig:histClass}
\end{figure}

For classification (see Figure~\ref{fig:histClass}), the principals are generally skewed towards high probabilities, indicating that the base models perform well and assign high probabilities to the correct class. An exception is the \textit{Higgs} dataset, where the distribution of principals is more symmetric and resembles a normal distribution. This is the only case where the logarithmic score performs poorly as a proxy for the quadratic score. This may be due to the strong penalisation of low-probability correct classes in the logarithmic score which may not be well suited for this dataset. 

\begin{figure}[H]
    \centering
    \resizebox{\columnwidth}{!}{%
    \addtolength{\tabcolsep}{-0.5em}
    \begin{tabular}{ccccc}
       
        \textbf{Abalone} & \textbf{Auction Verification} &  \textbf{BNG EchoMonths} & \textbf{California Housing} & \textbf{Infrared Thermography Temp.}\\
        \includegraphics[width=0.3\linewidth]{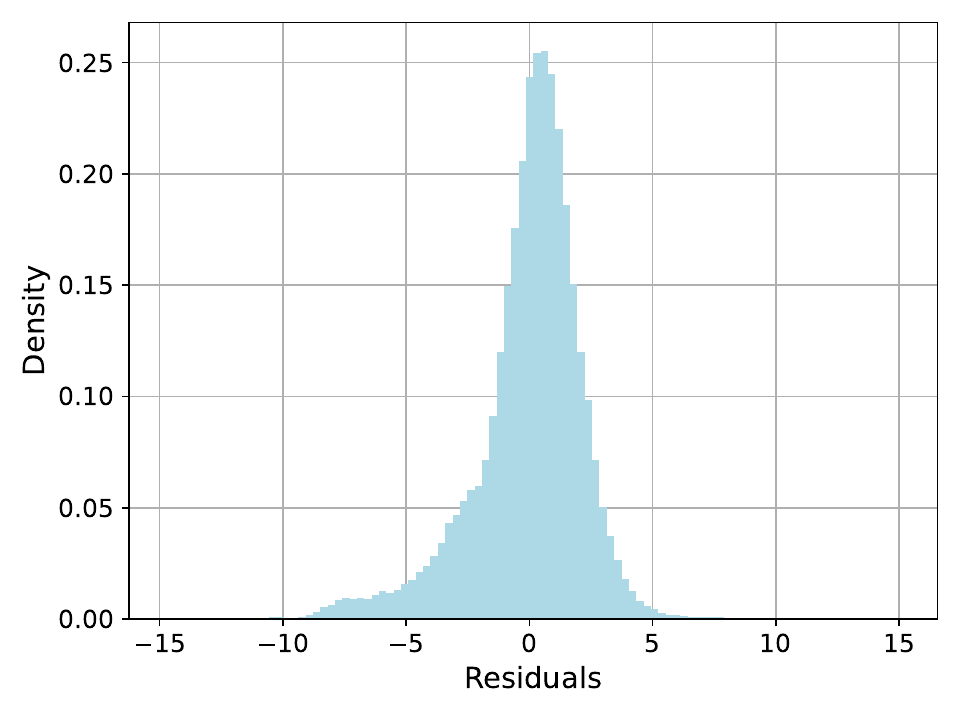} &
       \includegraphics[width=0.3\linewidth]{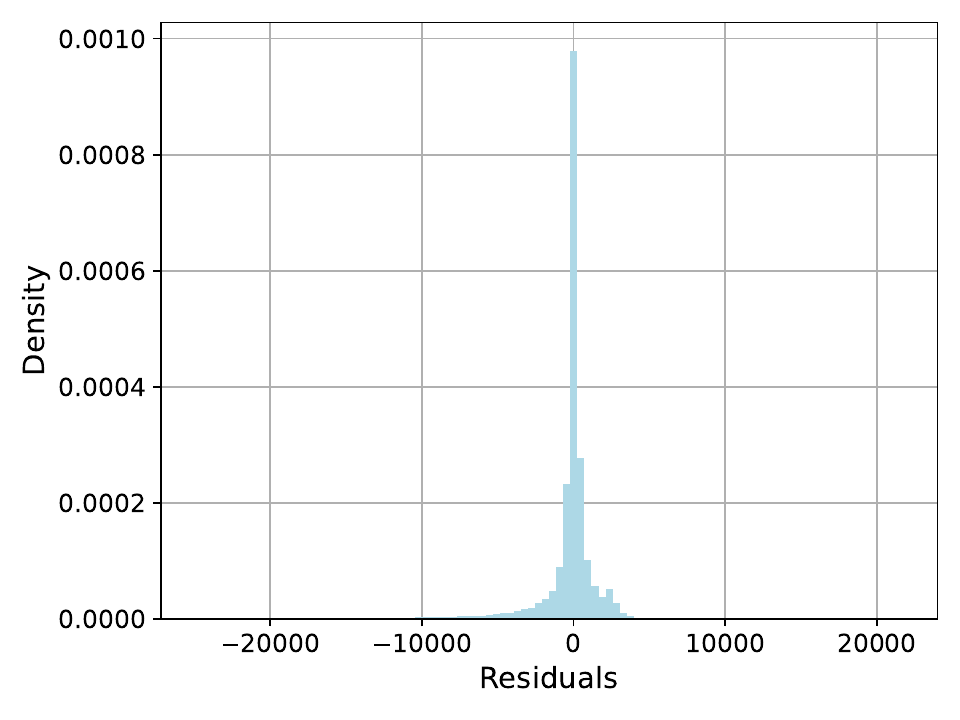} &
        \includegraphics[width=0.3\linewidth]{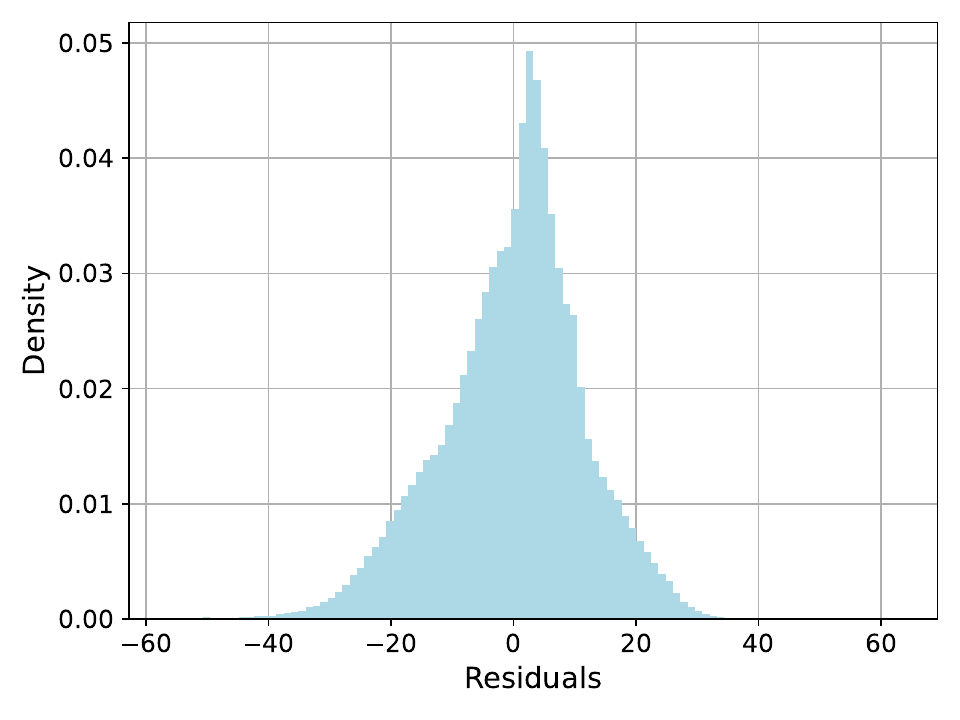} &
        \includegraphics[width=0.3\linewidth]{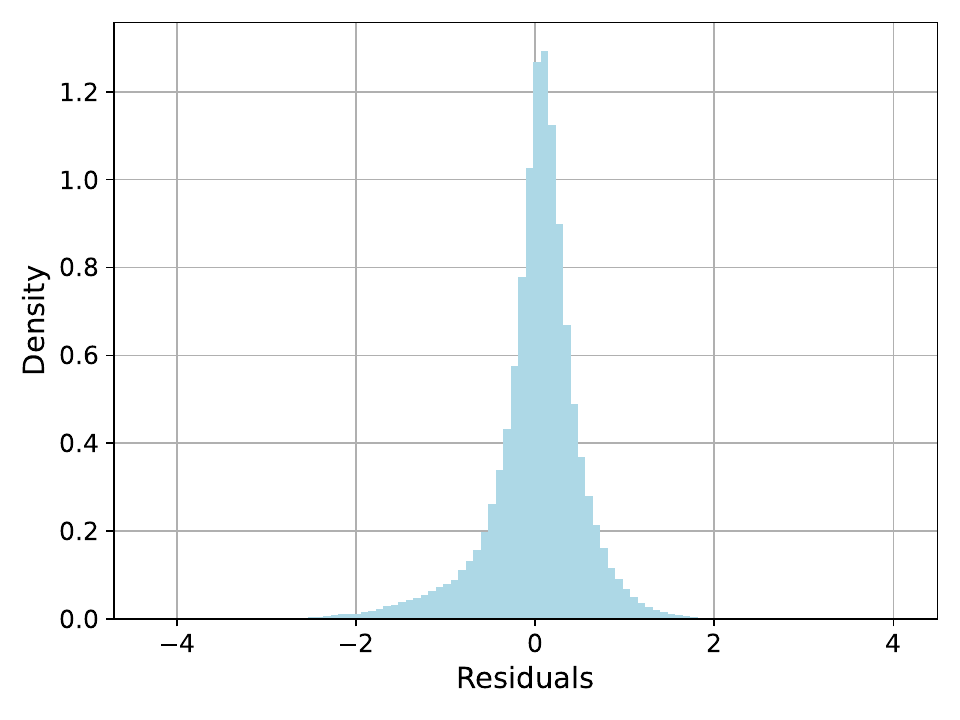} &
        \includegraphics[width=0.3\linewidth]{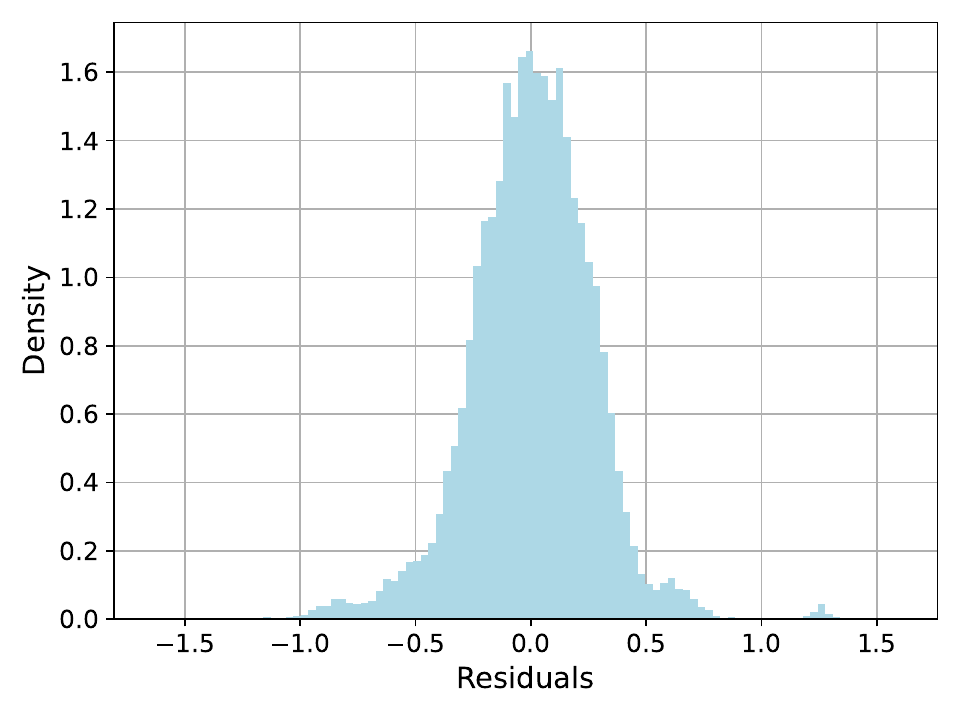}\\

         \textbf{Life Expectancy} &  \textbf{Music Popularity} & \textbf{Parkinsons Tele. (motor)} &  \textbf{Parkinsons Tele. (total)} & \textbf{Software Cost Estimation}\\
     
        \includegraphics[width=0.3\linewidth]{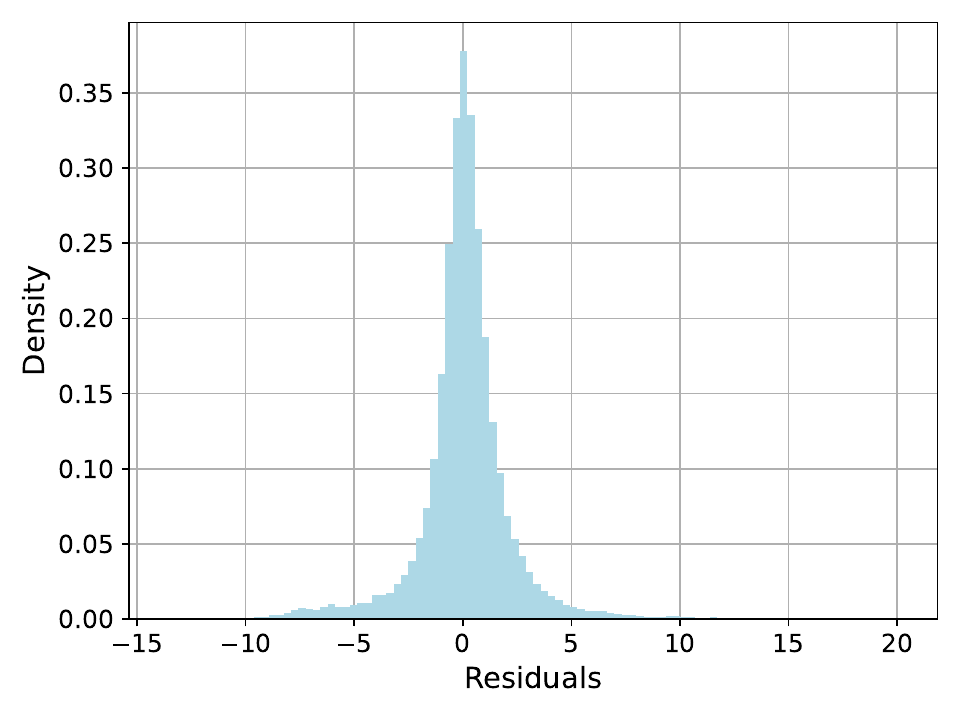} &       
        \includegraphics[width=0.3\linewidth]{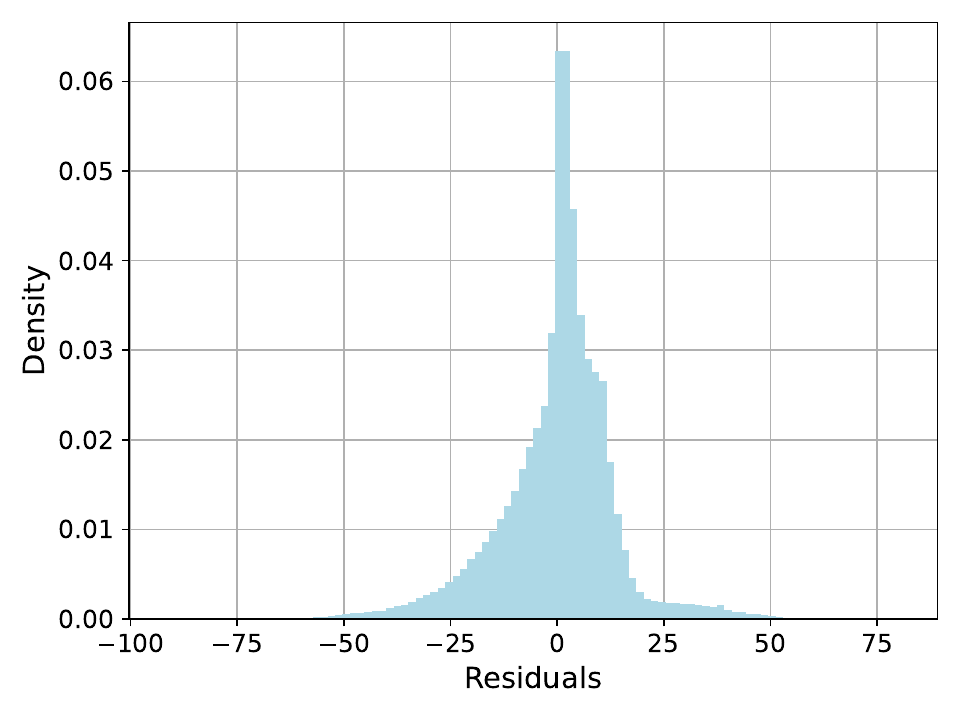} &
        \includegraphics[width=0.3\linewidth]{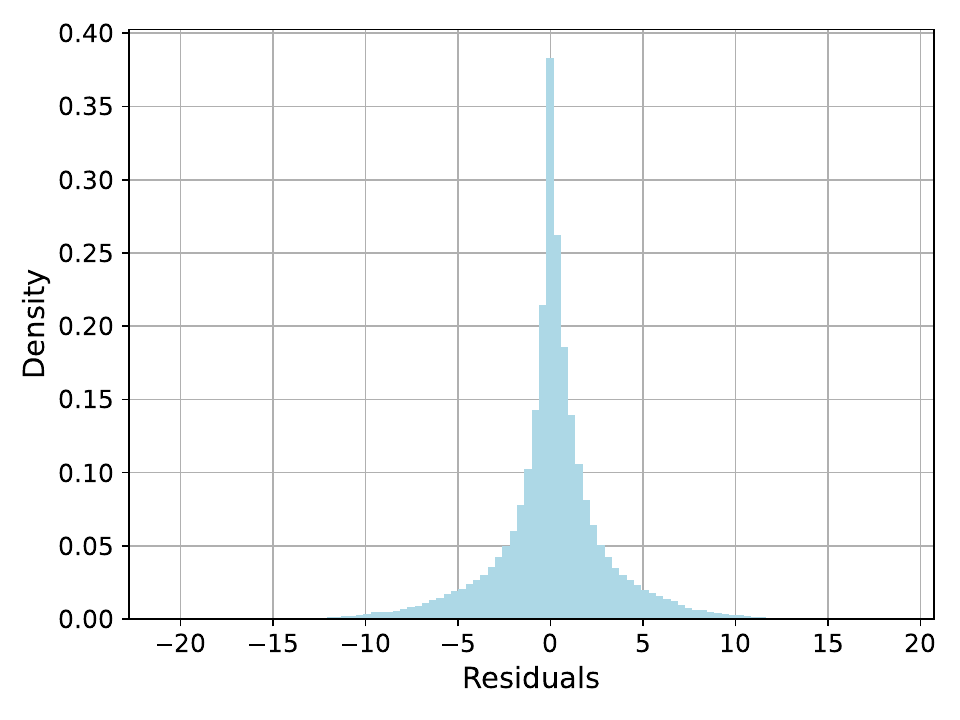} &
       \includegraphics[width=0.3\linewidth]{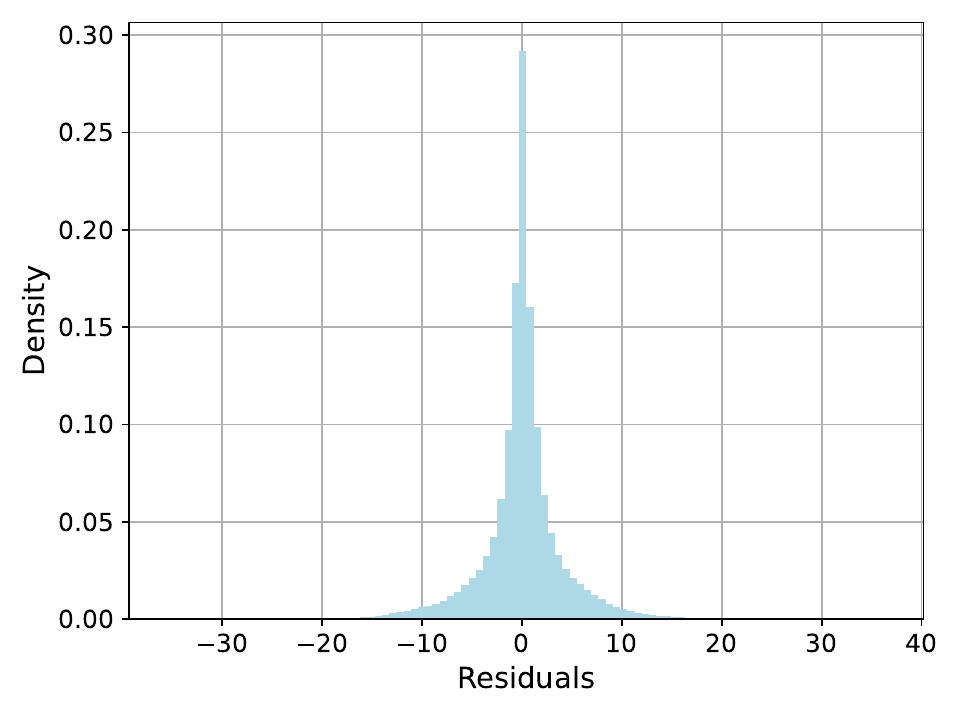} &
         \includegraphics[width=0.3\linewidth]{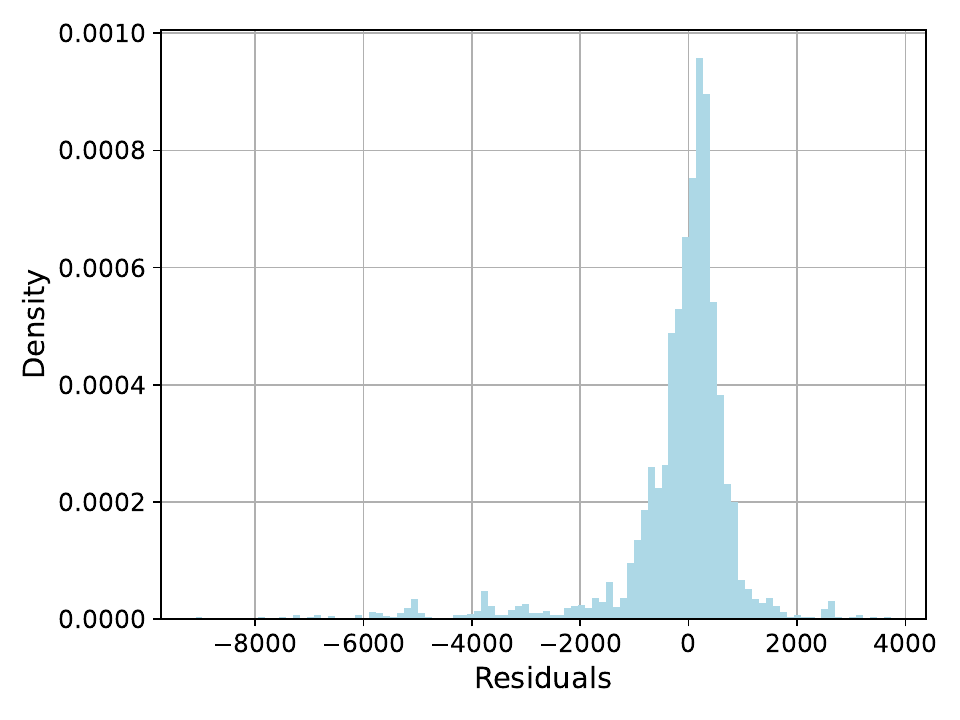}\\
    \end{tabular}
    }

    \caption{Histograms of residuals. Some datasets like \ds{Auction Verificatoin} have some very high residuals, which may contribute to the double penalisation on some losses.}
    \label{fig:histRegr}
\end{figure}

For regression  (see Figure~\ref{fig:histRegr}), in general, the residuals are centred around zero, indicating that on average the base models provide unbiased estimates. Exceptions, such as \ds{Auction Verification} have very high residuals, indicating that the base models occasionally make large prediction errors. This can contribute to the ``double punishment" phenomenon under certain loss functions, such as quadratic loss, penalise both the size of the residual and its square.

\newpage
\section{Score results for all assessor types (regression)}
\label{app:more_matrices}

\begin{figure}[H]
\vspace{-0.5cm}
    \centering
    \includegraphics[width=0.57\linewidth]{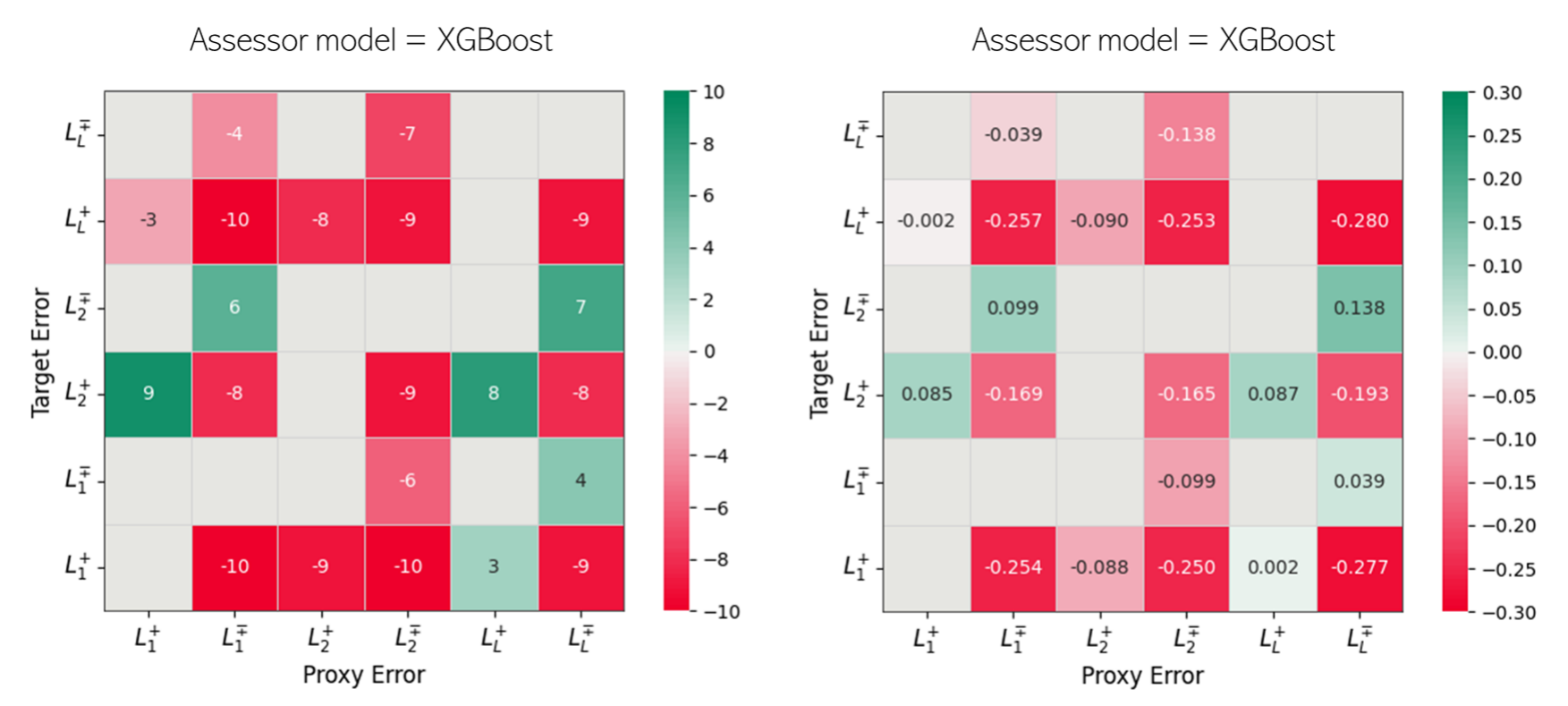}\vspace{-0.15cm}
        \includegraphics[width=0.57\linewidth]{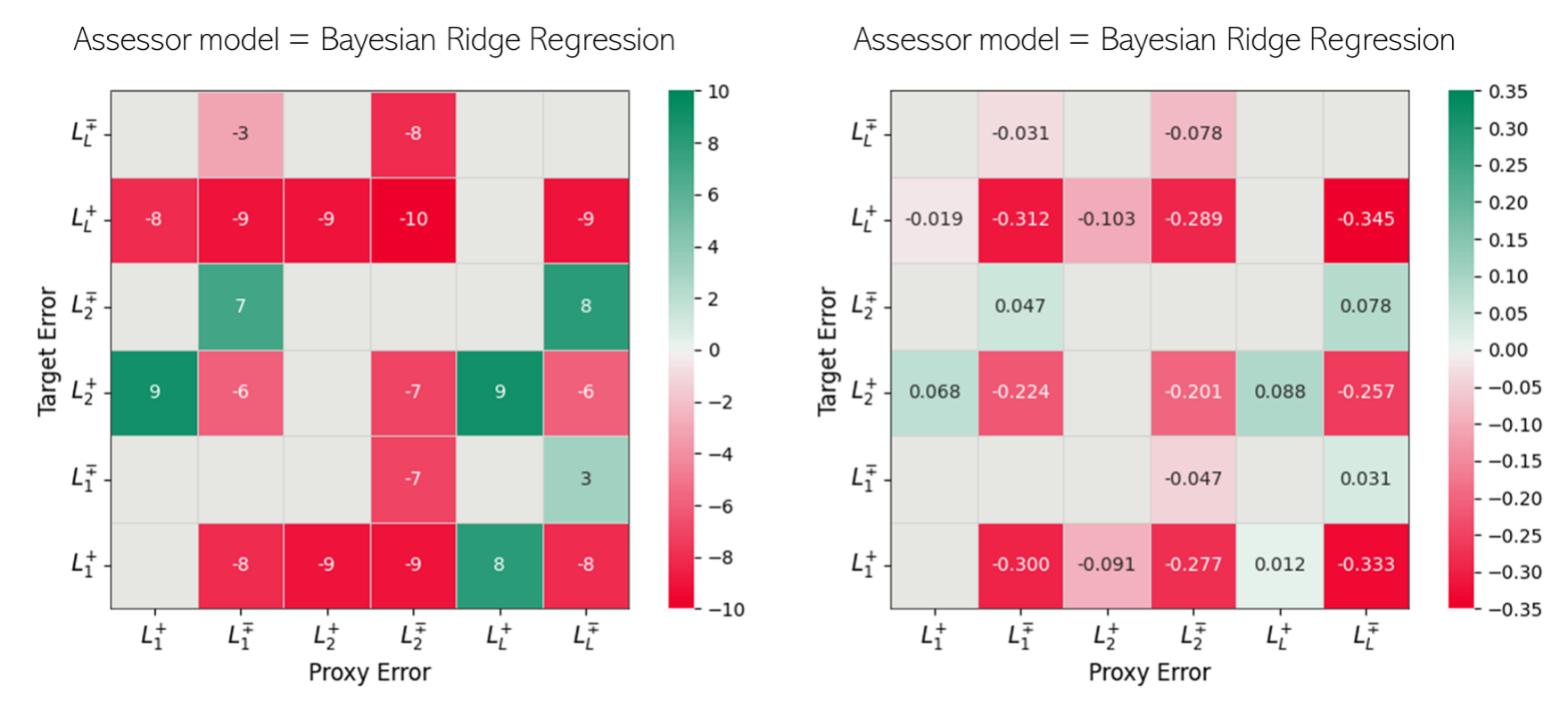}\vspace{-0.15cm}
            \includegraphics[width=0.57\linewidth]{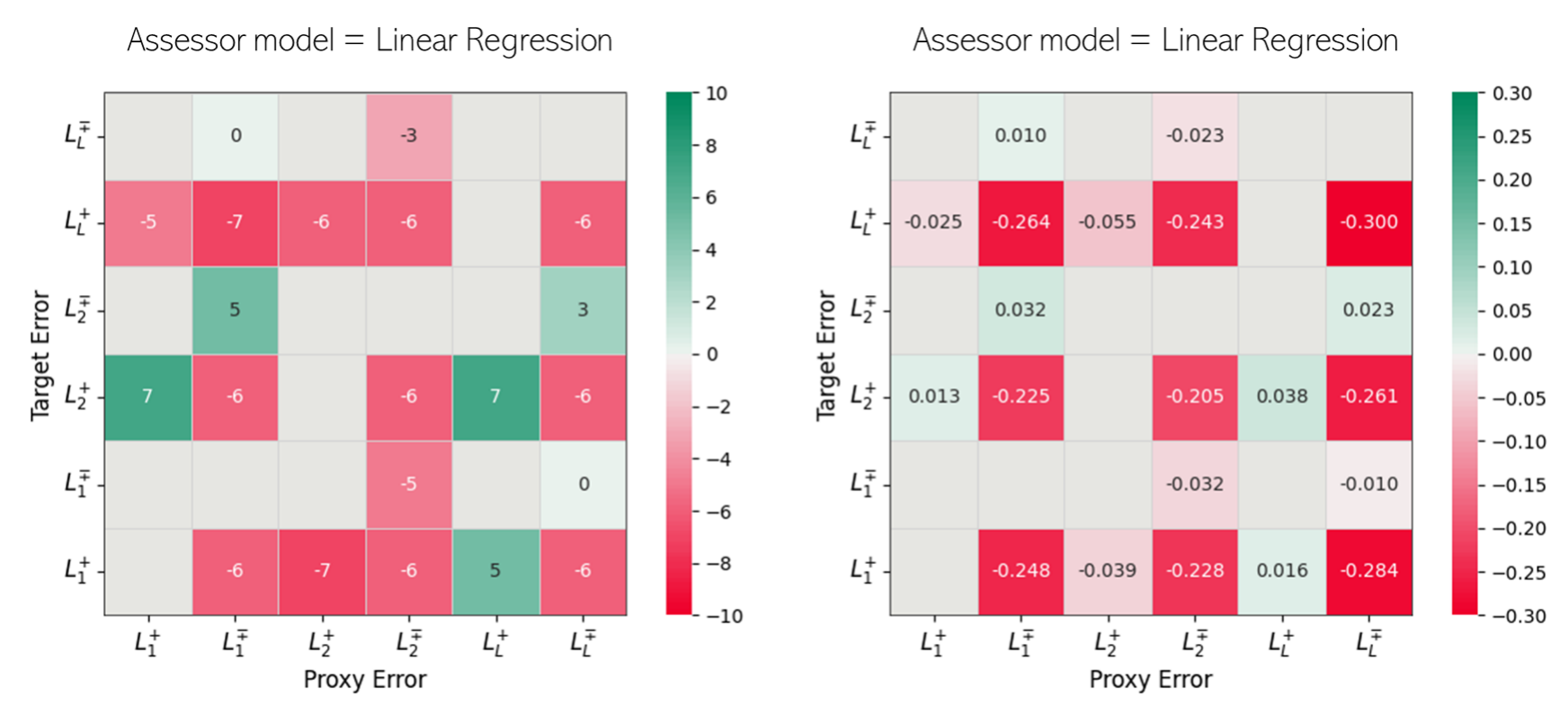}\vspace{-0.15cm}
                \includegraphics[width=0.57\linewidth]{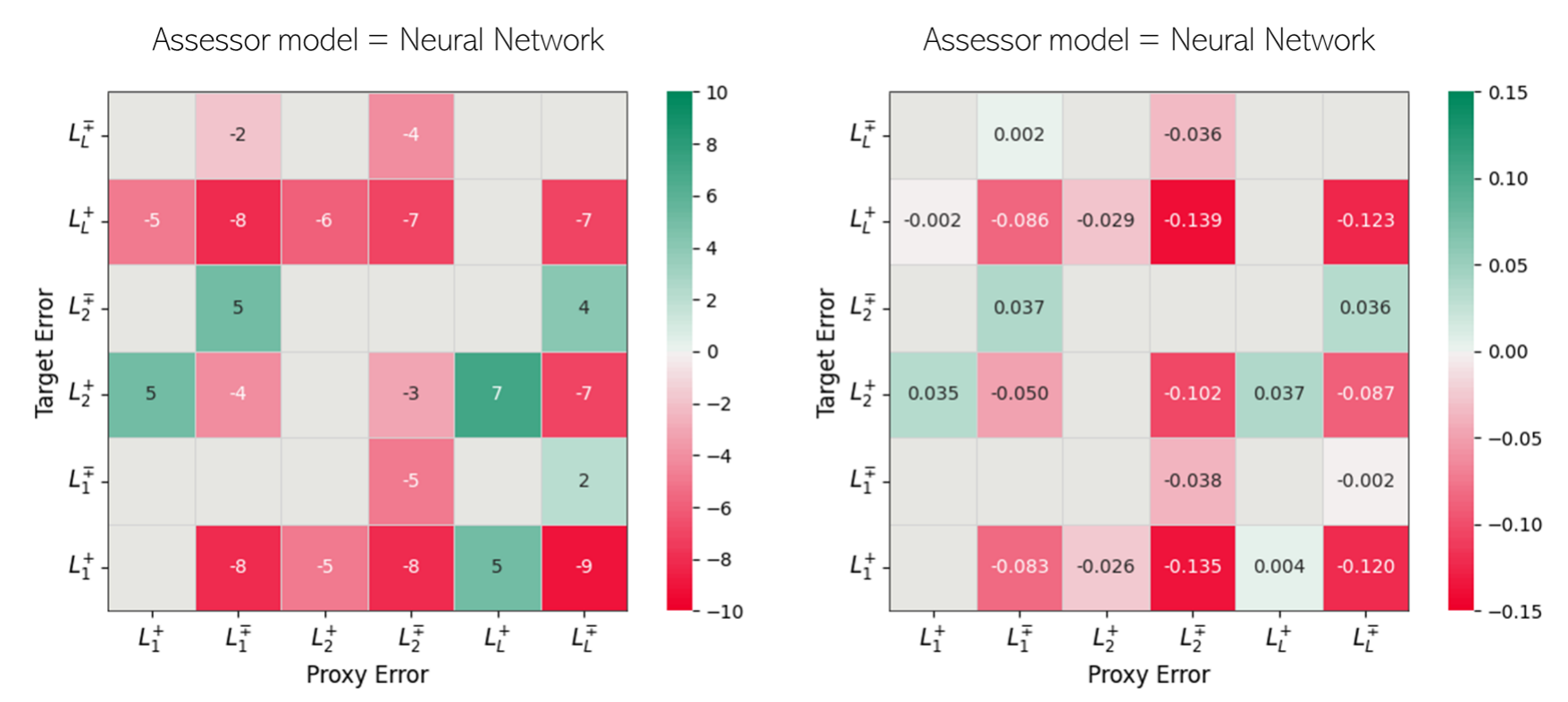}\vspace{-0.2cm}

    \caption{(Left) Score matrix for (by row) XGBoost, Bayesian ridge, Linear Regression and Feed-forward Neural Network assessor model. (Right) Aggregated Spearman margin matrix for XGBoost assessor model. In both matrices, rows represent target errors and columns proxy errors. Red values indicate poor performance from trying to predict $L_{\target}$ by learning $L_{\proxy}$. Inversely, green values show instances where learning from $L_{\proxy}$ is better than from learning directly from $L_{\target}$}
    \label{fig:app_all_score_matrix}
\end{figure}

This appendix provides detailed 
score and Spearman margin matrices for all the assessor models used in the regression tasks.  These results show that the findings discussed in the main text (using  XGBoost as the assessor) are consistent across different assessor models. Figure~\ref{fig:app_all_score_matrix} presents the score matrices for each assessor model. Each matrix shows the net score for using a proxy loss to predict a target loss across all datasets. Although the scores vary slightly (there are two groups with similar scores, XGBoost and Bayesian ridge regression vs Linear Regression and Neural Networks), the patterns observed are consistent across all assessor models: using signed losses as proxies for unsigned target losses generally results in poorer performance, often with negative scores. Also, the logistic errors prove to be successful proxies. 

\section{Underestimation of signed errors}\label{app:underpredictions}

\begin{figure}[H]
    \centering
    Proxy = \LabsN, Target = \LabsN\\
    
    \includegraphics[width=0.7\linewidth]{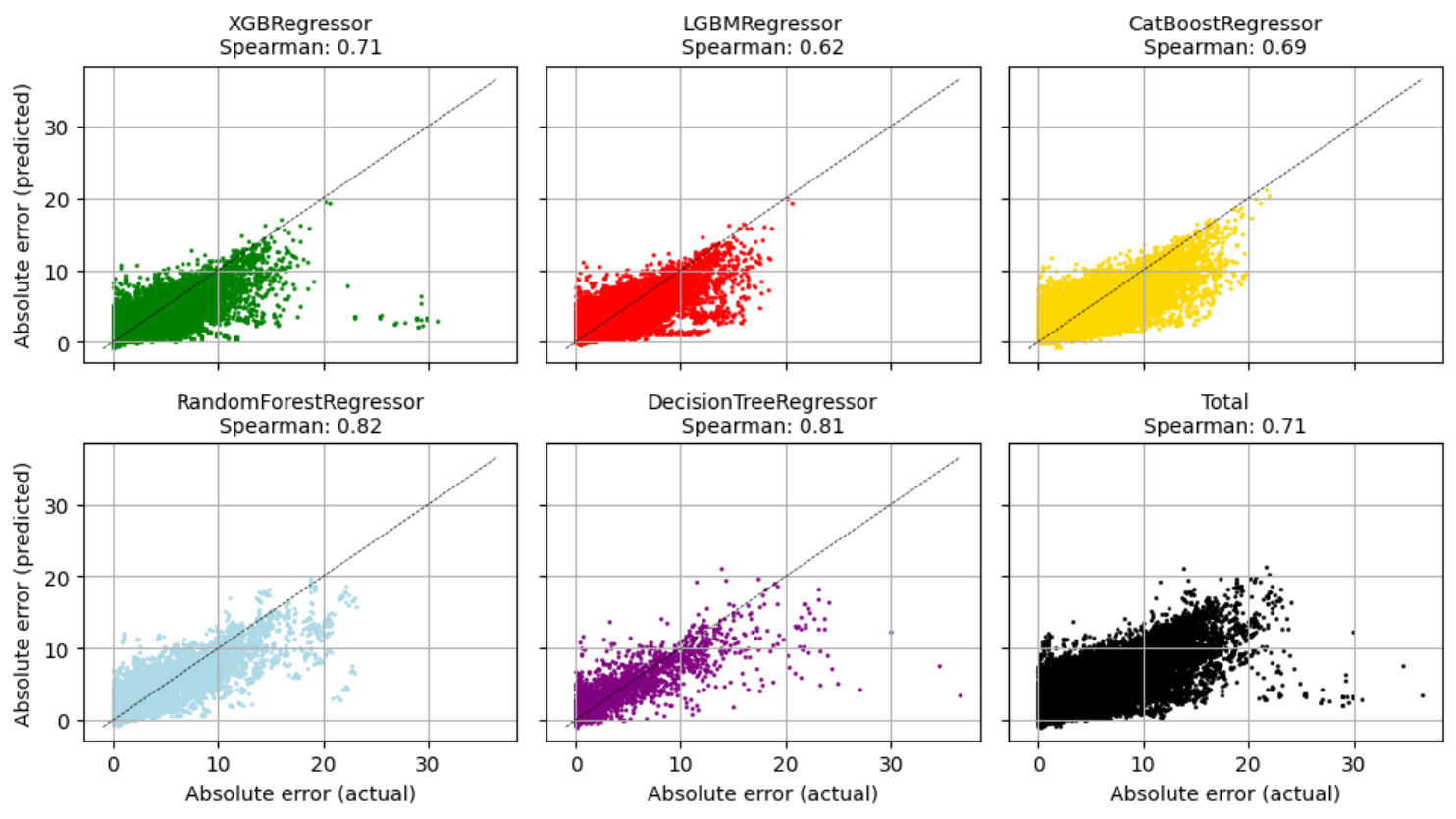}
    \\
    Proxy = \LsgnN, Target = \LabsN\\
    \includegraphics[width=0.7\linewidth]{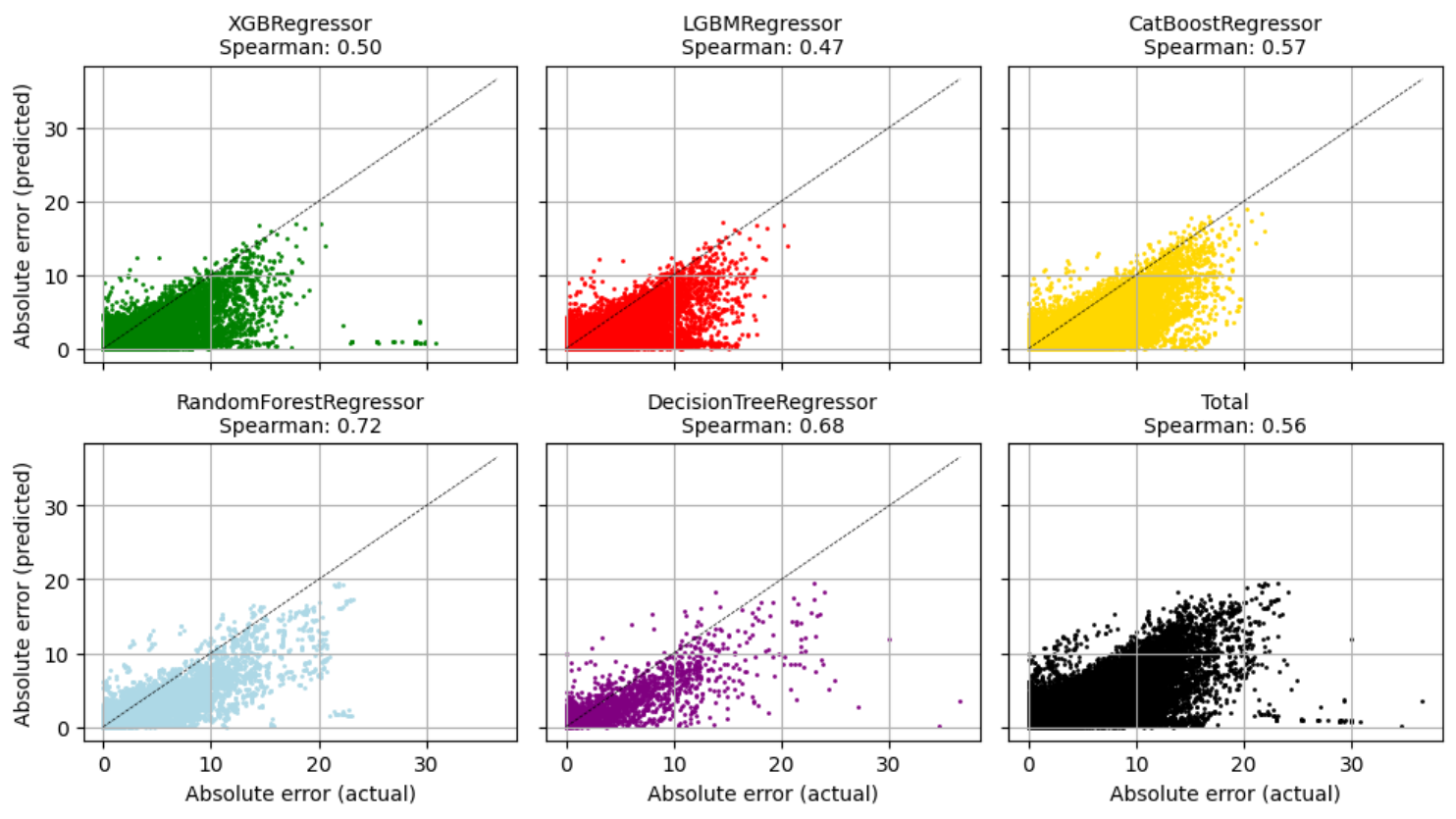}
    \caption{Scatter plots for the XGBoost assessor for the 
    Parkinson's Disease Rating Scale and five base models: XGBRegressor, LGBMRegressor, CatBoostRegressor, RandomForestRegressor and DecisionTreeRegressor. Because the predictions of the assessor tend to the mean, the case where the proxy is signed takes predictions towards 0, and the predictions usually fall under the diagonal. This behaviour appears in all base models}
    \label{fig:BigScatter}
\end{figure}

Here, following the discussion in the main text, we further analyse the phenomenon of underprediction observed when using signed errors as proxy losses to predict unsigned errors (specifically, Figure~\ref{fig:underprediction}). 
When assessors are trained using signed losses to predict unsigned target losses, we observe that the predicted losses tend to underestimate the true losses. This underestimation is evident in the scatter plots in Figure~\ref{fig:BigScatter}, where the predictions fall below the diagonal line (ideal prediction) for higher loss values.

The underestimation can be attributed to two main factors: mean reversion and loss of magnitude information. The underestimation of errors in signed losses may be due to two main factors: mean reversion and loss of magnitude information. Signed losses tend to average close to zero because positive and negative errors cancel each other out, causing assessor trained on these losses to predict values around zero and underestimate larger errors. In addition, while the sign indicates the direction of the error, it fails to convey magnitude when converted to an unsigned loss, leading to predicted values being compressed towards zero.

\end{document}